\documentclass{article}

\usepackage{natbib}
\usepackage{microtype}
\usepackage{graphicx} \graphicspath{{figures/}}
\usepackage{subfigure}
\usepackage{booktabs} 

\usepackage{amsmath}
\usepackage{amsfonts}       
\usepackage{nicefrac}      
\usepackage{tabularx}
\usepackage{multirow}
\usepackage{balance}
\usepackage{graphicx}
\usepackage[export]{adjustbox}

\usepackage{algorithm, algorithmic} 
\usepackage[algo2e]{algorithm2e}
\newcolumntype{?}{!{\vrule width 1pt}}

\usepackage{hyperref}

\usepackage[accepted]{icml2020}

\icmltitlerunning{Partly Supervised Multitask Learning}

\begin{document}

\twocolumn[
\icmltitle{Partly Supervised Multitask Learning}

\begin{icmlauthorlist}
\icmlauthor{Abdullah-Al-Zubaer Imran}{ed,to}
\icmlauthor{Chao Huang}{to}
\icmlauthor{Hui Tang}{to}
\icmlauthor{Wei Fan}{to}\\
\icmlauthor{Yuan Xiao}{goo}
\icmlauthor{Dingjun Hao}{goo}
\icmlauthor{Zhen Qian}{to}
\icmlauthor{Demetri Terzopoulos}{ed,foo}
\end{icmlauthorlist}

\icmlaffiliation{ed}{Department of Computer Science, University of California, Los Angeles, California, USA}
\icmlaffiliation{to}{Tencent Hippocrates Research Lab, Palo Alto, California, USA}
\icmlaffiliation{goo}{Xi’an Jiaotong University College of Medicine, China}
\icmlaffiliation{foo}{VoxelCloud, Inc., Los Angeles, California, USA}

\icmlcorrespondingauthor{Abdullah-Al-Zubaer Imran}{aimran@cs.ucla.edu}

\icmlkeywords{Semi-supervised learning, self-supervision, multitask learning, adversarial learning, spine x-ray, chest x-ray, segmentation, classification}

\vskip 0.3in
]

\printAffiliationsAndNotice{}  

\begin{abstract}
Semi-supervised learning has recently been attracting attention as an alternative to fully supervised models that require large pools of labeled data. Moreover, optimizing a model for multiple tasks can provide better generalizability than single-task learning. Leveraging self-supervision and adversarial training, we propose a novel general purpose semi-supervised, multiple-task model---namely, self-supervised, semi-supervised, multitask learning (S$^4$MTL)---for accomplishing two important tasks in medical imaging, segmentation and diagnostic classification. Experimental results on chest and spine X-ray datasets suggest that our S$^4$MTL model significantly outperforms semi-supervised single task, semi/fully-supervised multitask, and fully-supervised single task models, even with a 50\% reduction of class and segmentation labels. We hypothesize that our proposed model can be effective in tackling limited annotation problems for joint training, not only in medical imaging domains, but also for general-purpose vision tasks.
\end{abstract}

\section{Introduction}

The success of fully-supervised deep convolutional neural networks (CNNs) currently relies on large quantities of labeled data. However, in certain applications such as medical imaging, such annotated data are often either unavailable or expensive and time-consuming to obtain. Therefore, for real-world applications, there is growing interest in leveraging limited quantities of labeled data along with much greater quantities of unlabeled data. Such approaches are called semi-supervised learning (SSL). For them to be effective, however, the knowledge gained from the unlabeled data must be significant to the model \cite{chapelle2009semi}.

Depending on how unlabeled data are leveraged, semi-supervised learning can be accomplished in several ways, and this has recently emerged as a growing body of research, yielding schemes such as unsupervised domain adaptation \cite{zhang2019bridging}, self-supervised learning \cite{jing2019self}, adversarial learning \cite{donahue2016adversarial}, and multitask learning \cite{ruder2017overview}. In unsupervised domain adaptation, a model is pre-trained on similar tasks in some other domains with labeled data, and the pre-trained model is then fine-tuned with a limited set of labeled data in the target domain. Domain adaption can also be performed to learn a generic representation where the model is fully-supervised for source data and unsupervised for the target data \cite{yang2019domain, zhuang2019comprehensive}. Self-supervised learning is closely related to transfer learning. Unlike transfer learning, the model is pre-trained on some surrogate tasks in the same domain, and then the pre-trained model is evaluated on the main application tasks in computer vision or medical imaging. Self-supervised learning is usually based on the assumption that the predicted labels from the original data and the augmented data should be the same \cite{jing2019self}. Adversarial learning, like Generative Adversarial Networks (GANs) \cite{goodfellow2014generative}, augments the class labels with an additional label to differentiate the generated data and real data. A well balanced generator-discriminator helps towards learning useful visual features from the unlabeled data \cite{donahue2016adversarial}. Multitask learning (MTL) is basically defined as optimizing more than one loss. In MTL, multiple related tasks are jointly learned, which results in better generalization of the model \cite{ruder2017overview}. Moreover, jointly learning multiple tasks as a general objective improves performance compared to single-task learning \cite{zhang2017survey}. 

\begin{figure*}
    \centering
    \resizebox{0.9\linewidth}{!}{
    \begin{tabular}{l l}
    \includegraphics[width=\linewidth, trim={0.26cm 0.3cm 0.26cm 0cm},clip]{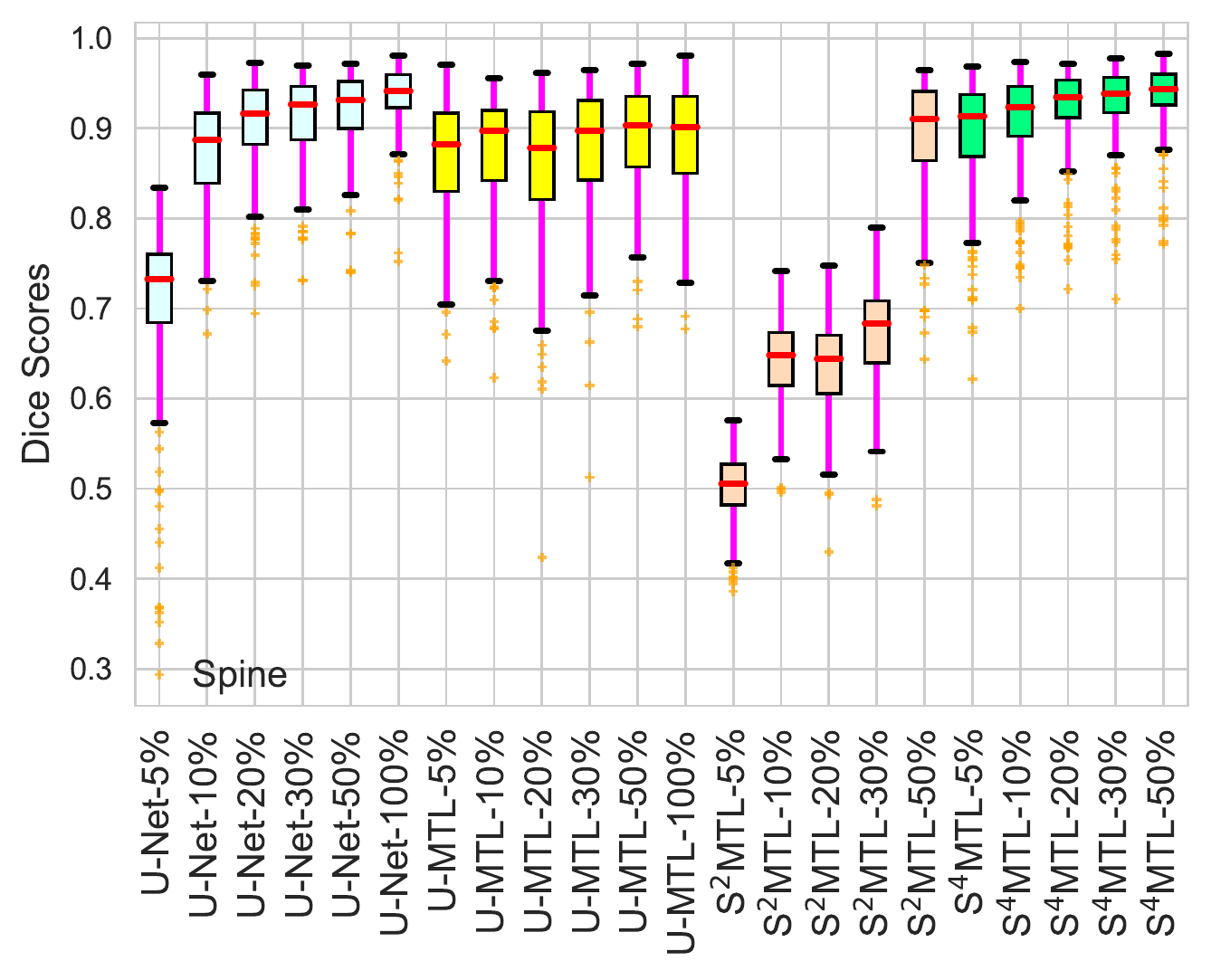}
    &
    \includegraphics[width=\linewidth, trim={0.26cm 0.3cm 0.26cm 0cm},clip]{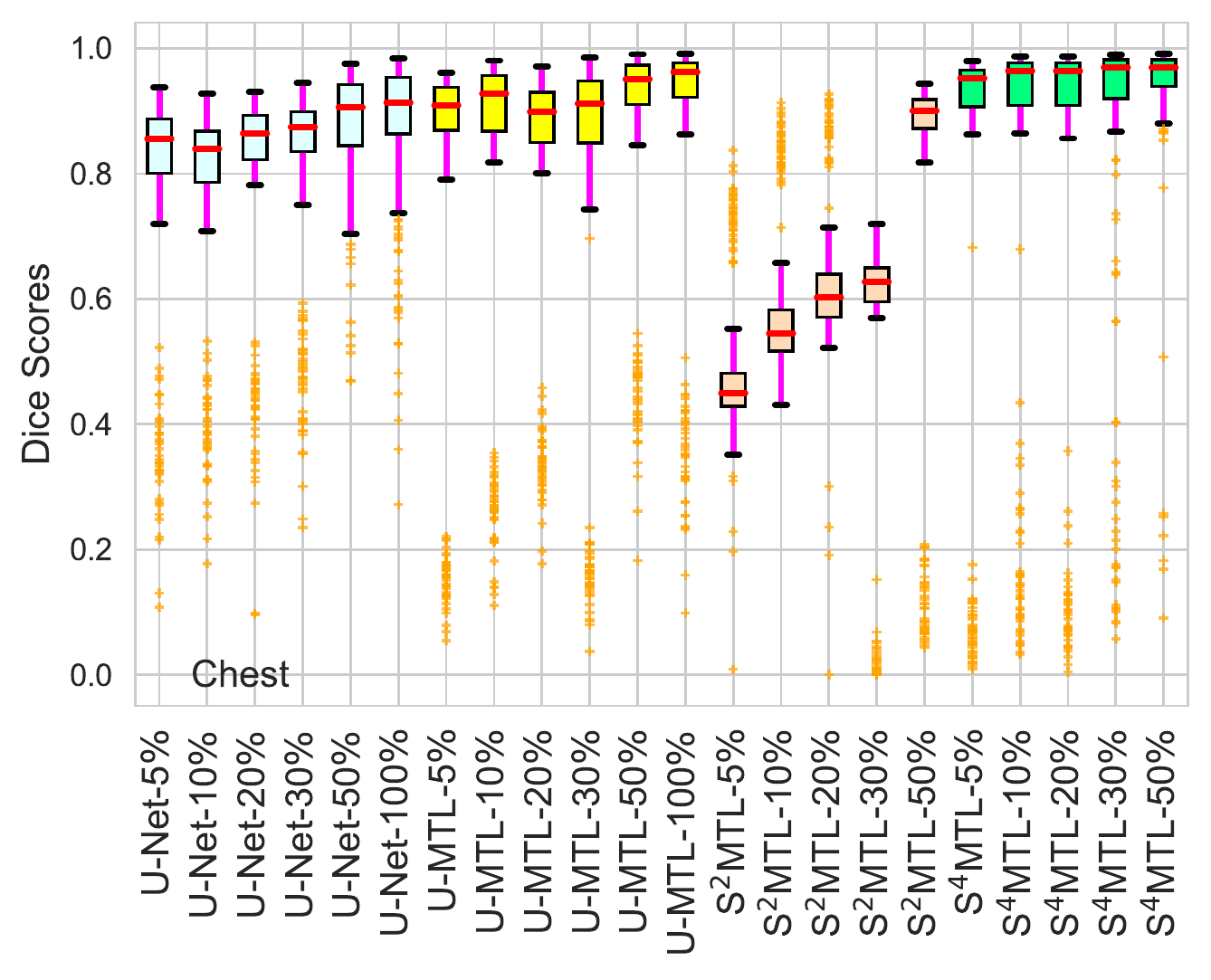}
    \end{tabular}
    }
    \caption{Box-Whisker plots showing consistent improvement in segmentation performance by the S${}^4$MTL model over baseline semi-supervised and fully-supervised models in varying proportions of labeled training data.}
    \label{fig:box_plots}
\end{figure*}

We propose a self-supervised, semi-supervised, multitask learning (S$^4$MTL) model that combines the advantages of self-supervised learning, adversarial learning, and multitask learning for real applications. Unlike simply optimizing multiple losses in the model, multiple real tasks are accomplished in our proposed approach. To demonstrate its effectiveness, our S$^4$MTL model is applied to two of the most important tasks in medical imaging---segmentation of anatomical structures and diagnostic classification---and both tasks are tackled by the same model. Clinically, it is important to label an image as normal or abnormal, but also to segment the relevant anatomical structures. Our model leverages a large proportion of unlabeled data to learn to accomplish both tasks, which distinguishes it from existing methods where the semi-supervision applies to only one task in a multitask learning setting \cite{zhou2019collaborative}. 

Our specific contributions can be summarized as follows:
\begin{itemize}
    \item A novel, general purpose, semi-supervised, multitask learning model leveraging self-supervision and adversarial training.
    \item The innovative use of chest and spine X-ray image data to tackle two important healthcare problems.
    \item Extensive experimentation showing the effectiveness of our model against state-of-the-art semi-supervised and fully-supervised single/multi-task models.
\end{itemize}

\section{Related Work}

We will now review the most relevant developments in computer vision and medical imaging in the field of semi-supervised learning and multitask learning.

\subsection{Semi-Supervised Learning}

Semi-supervised learning has recently been explored both in computer vision and medical imaging due to the availability of vast amounts of unlabeled data and computing power to process them. Semi-supervised learning is usually performed with a small portion of labeled and a larger portion of unlabeled data, assuming that both are from the same or similar distributions. In standard protocols, semi-supervised models are evaluated by retaining only a portion of the labels from a dataset while the remainder are treated as unlabeled data \cite{zhai2019s}. Depending on the approach to gaining information from large quantities of unlabeled data, semi-supervised learning can be performed in at least two different ways---self-supervised and adversarial training.

\begin{figure}
\centering
\resizebox{\linewidth}{!}{
\begin{tabular}{l l}
    \includegraphics[width=\linewidth, trim={0.8cm 0.5cm 1.2cm 0.7cm},clip]{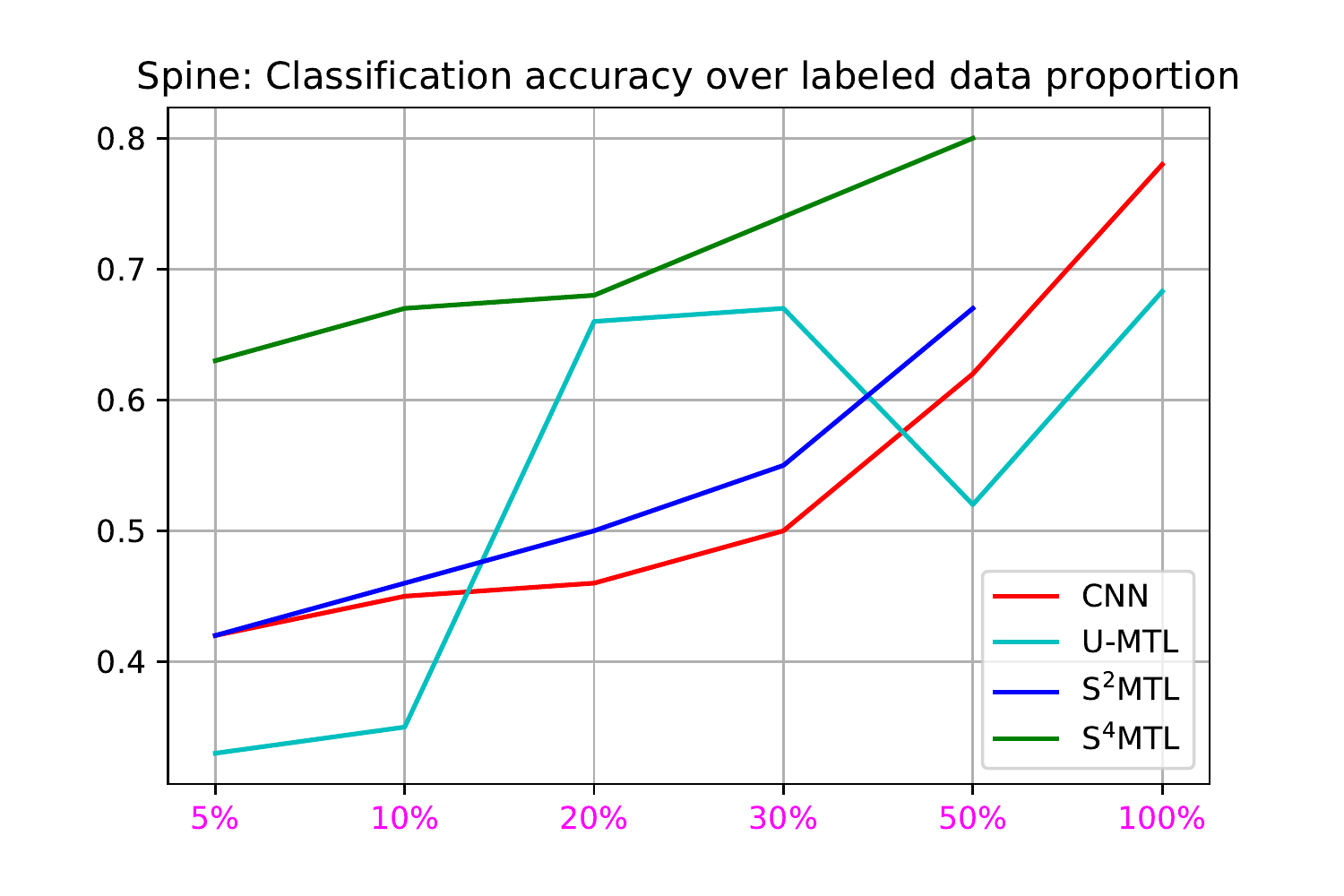}
    &
    \includegraphics[width=\linewidth, trim={0.8cm 0.5cm 1.2cm 0.7cm},clip]{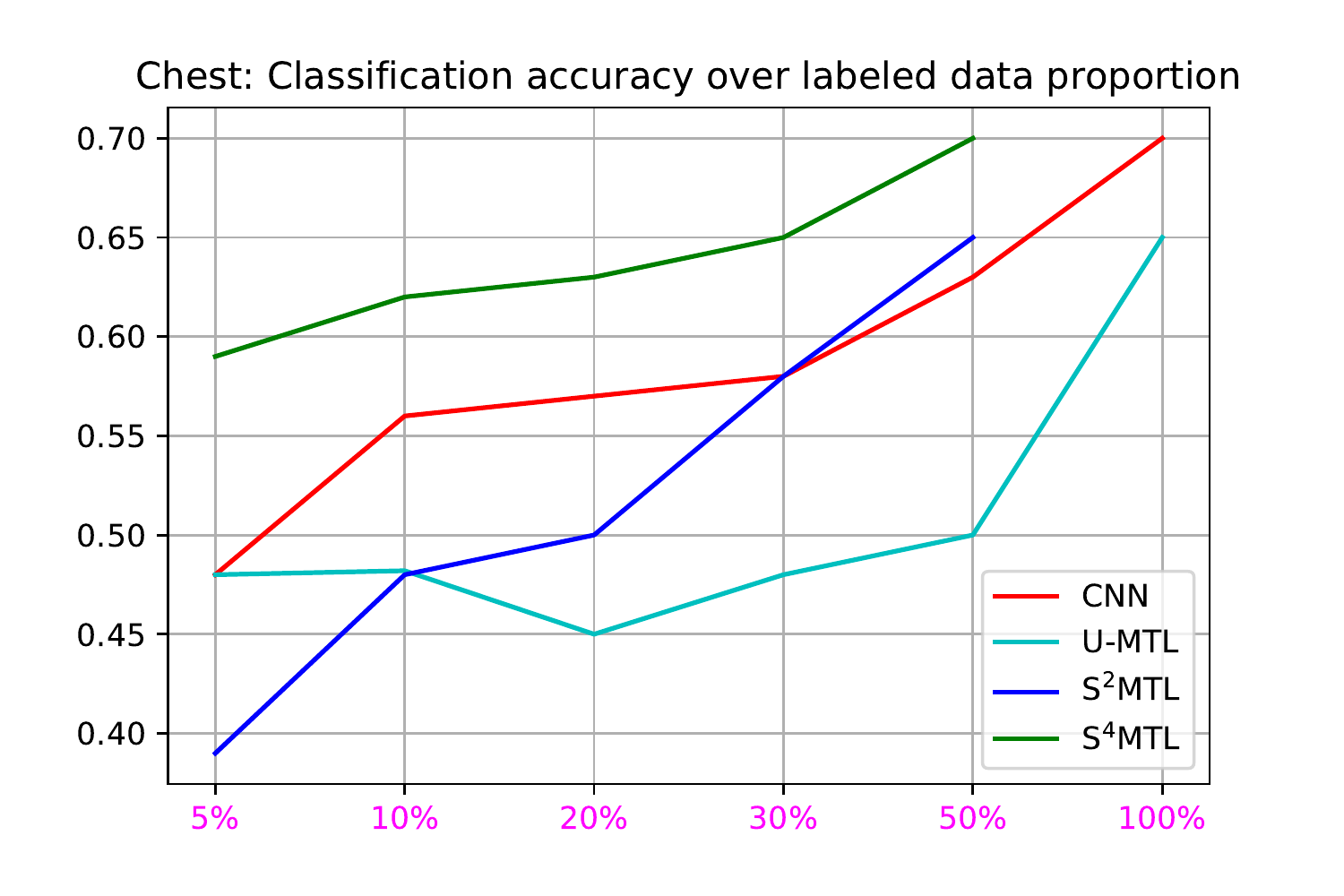}
    \end{tabular}
    }
    \caption{Consistent improvement in classification accuracy by the proposed S${}^4$MTL model over baseline semi-supervised and fully-supervised models in varying proportions of labeled training data.}
    \label{fig:class_plots}
\end{figure}

Self-supervised learning is similar to unsupervised learning in its goal of using a vast amount of unlabeled data to learn visual representation without any human annotation. Usually, self-supervised learning is performed by formulating a pretext or surrogate task only on the unsupervised data portion. Examples of pretext tasks could include image reconstruction, image colorization, predicting image rotations, etc. In self-supervision, data itself lends for supervision; i.e., proxy labels created from the data on which training can provide useful visual features from unlabeled data. \citet{tajbakhsh2019surrogate} showed the effectiveness of training models from pre-trained surrogate tasks in different medical imaging applications, including diabetic retinopathy classification, nodule detection, and lung lobe segmentation with limited labeled data. Moreover, without training separately, both the pretext and downstream tasks can be combined in jointly learning useful visual features. \citet{tran2019semi} proposed a semi-supervised learning scheme based on self-supervision, where the model is trained like full-supervision---a supervised branch for the labeled data and a self-supervised branch for the unlabeled data---to predict some geometric transformations. 

Adversarial learning can effectively be adapted to semi-supervised learning for classification of both natural and medical images \cite{salehinejad2018generalization, imran2019multi}. Adversarial learning has also been utilized in segmentation (semantic-aware generative adversarial nets \cite{chen2018semantic}, structure correcting adversarial nets \cite{dai2018scan}, etc.) as well as in disease classification (semi-supervised domain adaptation \cite{madani2018semi}, attention-guided CNN \cite{guan2018diagnose}).

\begin{figure}
    \centering
    \includegraphics[width=\linewidth, trim={2cm 5cm 3cm 4.5cm},clip]{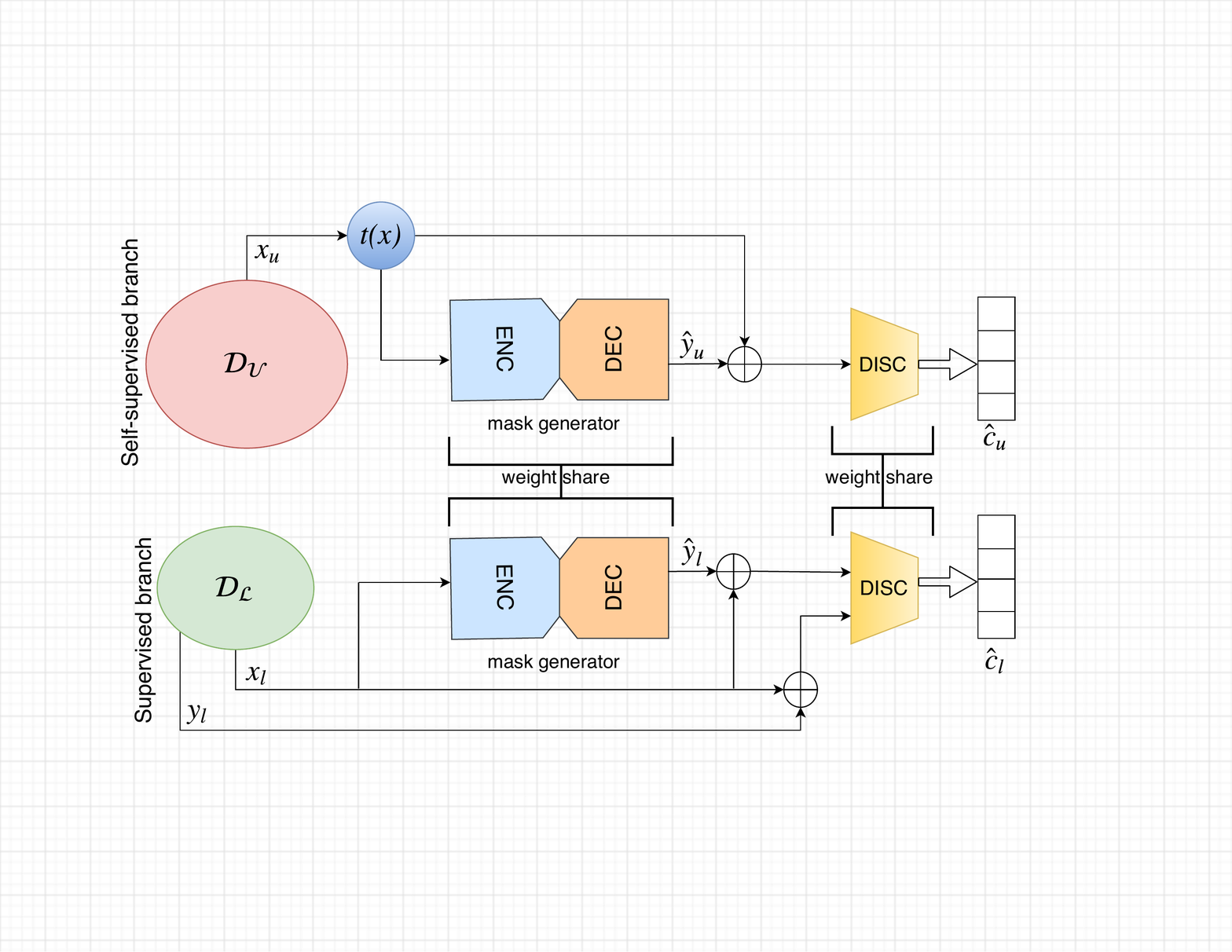}
    \caption{Schematic of the self-supervised, semi-supervised, multitask learning model (S$^4$MTL). A segmentation mask generator produces masks on taking inputs from labeled or unlabeled samples. A class discriminator takes concatenated inputs from labeled data-mask or unlabeled data-mask pairs and predicts the class labels. For the labeled data branch, it is fully supervised. By contrast, using unlabeled data, the self-supervised branch employs self-generated labels using a geometric transformation function $t(x)$. The predicted segmentation output is obtained from the decoder (DEC) and the diagnostic classification prediction is received at the discriminator (DISC).}
    \label{fig:arch}
\end{figure}

\subsection{Multitask Learning}

Multitask learning (MTL) can be accomplished in several ways, such as learning from auxiliary tasks to support the main task \cite{liebel2018auxiliary}, learning to learn multitask models \cite{zhang2018learning}, joint learning of multiple tasks \cite{liu2019joint, imran2019semi}, etc. By performing multiple tasks, the domain-specific information in the training signals of related tasks is actually improved \cite{Caruana1993MultitaskLA}. MTL is particularly useful for implicit data augmentation through better representation, focusing attention on the most relevant features, learning one task through another, and regularizing among them \cite{ruder2017overview}. 

The literature includes several efforts on performing multiple tasks within the same model. Several prior efforts address multitask learning with CNNs and generative modeling. \citet{Rezaei2018MultiTaskGA} combined a set of auto-encoders with an LSTM unit and an FCN as discriminator for semantic segmentation and disease prediction.  \citet{girard2019joint} used a U-Net-like architecture coupled with graph propagation to jointly segment and classify retinal vessels. \citet{Mehta2018YNetJS} proposed a Y-Net, with parallel discriminative and convolutional modularity, for the joint segmentation and classification of breast biopsy images. Another multitasking model was proposed by \citet{yang2017novel} for skin lesion segmentation and melanoma-seborrheic keratosis classification, using GoogleNet extended to three branches for segmentation and two classification predictions. \citet{khosravan2018semi} used a semi-supervised multitask model for the joint learning of false positive reduction and nodule segmentation from 3D computed tomography (CT) images. Most recently, \citet{imran2019semi} proposed an SSL model for the joint classification and segmentation of medical images in an adversarial training framework; however, no assumption was made on the distributions of labeled and unlabeled data and training was performed by simply masking out the labels for the unsupervised data.

As a departure from the existing SSL and MTL models, we propose the joint classification and segmentation of medical images via a novel SSL model with MTL leveraged by self-supervision and adversarial learning.

\section{The S$^{\bf 4}$MTL Model}
\label{sec:methods}

We combine the concepts of self-supervision and adversarial learning in a semi-supervised multitask learning scheme for jointly performing image recognition and semantic segmentation within the same model. To formulate the problem, we assume an unknown data distribution $p(X,Y,C)$ over images, segmentation labels, and class labels. The model has access to the labeled training set $\mathcal{D_L}$ sampled i.i.d.~from $p(X, Y, C)$ and to the unlabeled training set $\mathcal{D_U}$ sampled i.i.d.~from $p(X)$ after marginalizing out $Y$ and $C$. We set the learning objectives for both classification and segmentation tasks as
\begin{equation}
    \label{eqn:main-obj}
    \min_{\psi, \theta}\ \mathcal{L_L}(\mathcal{D_L}, (\psi, \theta)) + \alpha\mathcal{L_U}(\mathcal{D_U}, (\psi, \theta)),
\end{equation}
where the supervised loss $\mathcal{L_L}$ is defined on the labeled data and the unsupervised loss $\mathcal{L_U}$ is defined on the unlabeled data, $\alpha$ is a non-negative weight parameter, and $\psi$ and $\theta$ denote the parameters of the classification and the segmentation networks, respectively.

 Referring to Fig.~\ref{fig:arch}, the S${}^4$MTL model comprises two main components---a segmentation mask generator $G$ and a class discriminator $D$. Generator $G$ can be any segmentation network, such as U-Net \cite{ronneberger2015u}, and any CNN classifier \cite{lecun2010convolutional} may be used as discriminator $D$. The model has a supervised branch for the labeled data and a self-supervision branch for the unlabeled data, and the two branches share the same $G$ and $D$. Labeled data are passed through the supervised branch and supervised losses are calculated. Unsupervised losses are computed by feeding the unlabeled data through the unsupervised branch. The two networks $G$ and $D$ are trained in an adversarial learning manner, where the mask generator and the class discriminator compete against each other. The objective in (\ref{eqn:main-obj}) is therefore specified as two losses $\mathcal{L}_D$ and $\mathcal{L}_G$ for the two networks $D$ and $G$, respectively:
\begin{equation}
    \label{eq:obj2}
    \begin{aligned}
    & \min_\psi\ \mathcal{L}_D ( \mathcal{L}(\mathcal{D_L}, \psi_D) + \alpha\mathcal{L}(\mathcal{D_U}, \psi_D) ),\\ 
    & \min_\theta\ \mathcal{L}_G (\mathcal{L}(\mathcal{D_L}, \theta_G) + \alpha\mathcal{L}(\mathcal{D_U}, \theta_G)).    
\end{aligned}
\end{equation}
Algorithm~\ref{alg:train} specifies the overall training procedure of the S${}^4$MTL model.

\begin{algorithm}[t]
\caption{S$^4$MTL Mini-Batch Training.}
\label{alg:train}
\begin{algorithmic}
\REQUIRE 
\STATE Training set of labeled data $x_l, y_l, c_l \in \mathcal{D_L}$ 
\STATE Training set of unlabeled inputs $x_u \in \mathcal{D_U}$ 
\STATE Transformation function $t(x)$ to generate $c_u$ from $x_u$ 
\STATE Network architecture $D_\psi, G_\theta \in \mathcal F_{(\psi,\theta)}$ with learnable parameters $\psi$, $\theta$
\vspace{4pt}
\FOR{each epoch over $\mathcal{D_U}$}
\STATE Generate minibatches of unlabeled inputs $\mathcal{M_U}$ using $t(x)$
\vspace{4pt}
\FOR{each step}
\STATE Sample minibatch $x_{l_{(i)}};{x_{l_{(1)}},\dots,x_{l_{(m)}}} \sim p_{\mathcal{D_L}(x)}$
\STATE Sample minibatch $x_{u_{(i)}};{x_{u_{(1)}},\dots,x_{u_{(m)}}} \sim p_{\mathcal{D_U}(x)}$

\vspace{4pt}
Compute model outputs for the labeled inputs:\\ $\hat{y}_l,\hat{c}_l\leftarrow \mathcal F_{(\psi,\theta)}(\mathcal{M_L})$\\
Compute model outputs for the unlabeled inputs:\\ $\hat{y}_u,\hat{c}_u\leftarrow \mathcal F_{(\psi,\theta)}(\mathcal{M_U})$

\vspace{4pt}
\STATE Update the class discriminator $D$ along its gradient:
\begin{align*}
    \nabla_{\psi_{D}} &\frac{1}{|\mathcal{M_L}|}\sum_{i\in\mathcal{M_L}}\left[L_{D_{\left(x_{l_{(i)}}, y_{l_{(i)}}, \hat{y}_{l_{(i)}}, c_{l_{(i)}}, \hat{c}_{l_{(i)}}\right)}}\right] + \\
    &\alpha \frac{1}{|\mathcal{M_U}|}\sum_{i\in\mathcal{M_U}}\left[L_{D_{\left(x_{u_{(i)}},  \hat{y}_{u_{(i)}}, \hat{c}_{u_{(i)}}\right)}}\right]
\end{align*}

\STATE Update the segmentation mask generator $G$ along its gradient:
\begin{align*} 
\nabla_{\theta_{G}} &\frac{1}{|\mathcal{M_L}|}\sum_{i\in\mathcal{M_L}}\left[L_{S_{\left(x_{l_{(i)}}, y_{l_{(i)}}, \hat{y}_{l_{(i)}}\right)}}\right] + \\
    &\alpha \frac{1}{|\mathcal{M_U}|}\sum_{i\in\mathcal{M_U}}\left[L_{G_{\left(x_{u_{(i)}},  \hat{y}_{u_{(i)}}\right)}}\right]
\end{align*}
\ENDFOR
\ENDFOR
\end{algorithmic}
\end{algorithm}

\subsection{Self-Supervision}

Self-supervision is usually formulated on two tasks---a surrogate or pretext task and a downstream or main task. Unlike fine-tuning on the downstream task using the pre-trained model from the pretext task, we combine them and refer to the combination as the \emph{pre-stream} (i.e., {\it Pre}text $+$ down{\it stream}) task for concurrently performing supervision and self-supervision. Although the self-supervision can be applied to both labeled and unlabeled data, we confine self-supervision only to the unlabeled data. We define pretext tasks for the unlabeled data $\mathcal{D_U}$. The self-supervision applies to the class discriminator $D$, whereas it is still unsupervised at the mask generator $G$. For the classification, we use a transformation function $t(x)$ to randomly flip (horizontal/vertical) or rotate (0, 90, 180, etc.) the unlabeled images and allow the network $D$ predict them.    

\subsection{Classification}

The real samples and labels to $G$ are presented in the forward pass. In the backward pass, the feedback from $D$ is passed to $G$. In the original image generator GAN, the discriminator works as a binary classifier, classifying the input image as real or synthetic. In order to facilitate the training of an $n$-class classifier, we expand the role of our class discriminator $D$ to an $(n+1)$-classifier. For multiple logit generation, we replace the sigmoid function by a softmax function, such that it can receive concatenated image-mask $(x_l, y_l)$, image-predicted mask $(x_l, \hat{y_l})$ for labeled data, and $(x_u,\hat{y}_u)$ for unlabeled data as inputs, and output an $(n+1)$-dimensional vector of logits $\{{l}_1, {l}_2,\dots,{l}_{n+1}\}$, which are finally transformed into class probabilities for the final classification. The class probabilities for the labeled data are calculated as
\begin{equation}
\label{eqn:class_l}  
    p(\hat{c}_l = (c_l = i) | (x_l, y_l)) = \frac{\exp(l_{i})}{\sum_{j=1}^{n+1}\exp(l_j)}
\end{equation}
and for the unlabeled data as
\begin{equation}
\label{eqn:class_u}  
    p(\hat{c_u} = (c_u = i) | (x_u,\hat{y}_u)) = \frac{\exp(l_i)}{\sum_{j=1}^{n+1}\exp(l_j)}.
\end{equation}

\subsection{Segmentation}

The segmentation mask generator takes input $x_l$ and generates $\hat{y}_l$ for the labeled data. For the unlabeled data, mask prediction $\hat{y}_u$ is generated from input $x_u$. For the labeled data, it is just like regular supervised segmentation. We employ Dice loss for the base model. Since the ground truth mask $y_l$ is also available, the segmentation loss is calculated as 
\begin{equation}
    \label{eqn:seg_l}
    \begin{aligned}
    &L_{G_\mathcal{L}{(\textrm{seg})}} = 1 - \\ 
    &\qquad\frac{
  \sum_i^{m^2} y_{pk}^{(i)}\hat{y}_{pk}^{(i)}}{
  \sum_i^{m^2}y_{pk}^{(i)}\hat{y}_{pk}^{(i)} + 
    \frac{1}{2}\sum_i^{m^2} y_{p\bar{k}}^{(i)}\hat{y}_{pk}^{(i)} + \frac{1}{2}\sum_i^{m^2} y_{pk}^{(i)}\hat{y}_{p\bar{k}^{(i)}} 
  }.
  \end{aligned}
\end{equation}
On the other hand, the ground truth mask for the unlabeled data is not available; therefore, we cannot directly calculate the segmentation loss on the predicted mask $\hat{y}_u$. Instead, we use an unsupervised loss---logit-wise distribution matching via KL divergence. The logit-wise absolute KL divergence \cite{imran2019semi} between the ground truth of labeled data $y_l$ and prediction on unlabeled data $\hat{y}_u$ over each pixel $i$ and logit $k$ is 
\begin{equation}
    \label{eqn:seg_u_kl}
    L_{G_{\textrm{KL}}(\mathcal{D}_U)} = \sum_i^{m^2}\left| (y_{l_{pk}}{(i)} - \hat{y}_{u_{pk}}{(i)})\log({y_{l_{pk}}{(i)}}/{\hat{y}_{u_{pk}}{(i)}})\right|.
\end{equation}

\subsection{Final Losses}

Depending on the source of the model inputs, $G$ and $D$ have multiple objectives for labeled and unlabeled data, combined for training with gradient descent. 

Like any supervised learning model, $G$'s supervised loss is just based on the labeled samples (at pixel-level). We employ the generalized Dice loss in this regard. As in adversarial  training, the generator's objective includes segmentation loss and adversarial prediction loss, where the segmentation mask generator $G$ wants the class discriminator $D$ to maximize the likelihood for the generated segmentation masks. For the labeled examples, we calculate two-way losses from image-label and image-prediction pairs, which differs from the unlabeled examples, where only image-prediction pairs are taken into account. The segmentation loss terms are calculated using (\ref{eqn:seg_l})--(\ref{eqn:seg_u_kl}).  
The unsupervised adversarial prediction loss terms include adversarial prediction losses for the labeled and unlabeled data. The mask generator $G$ wants the class discriminator to maximize the likelihood for the image-prediction pairs $x_l, \hat{y}_l$. Therefore, the adversarial prediction loss in $G$ is
\begin{equation}
\label{eqn:G_pred_l}  
    L_{G_{{\textrm{pred}_{(x_l,\hat{y}_l)}}}} = - \mathbb{E}_{x_l,\hat{y}_l \sim G}\log[1 - p(c_l = n+1 | (x_l,\hat{y}_l)].
\end{equation} 
Similarly, for the unlabeled data, $x_u, \hat{y}_u$, the adversarial prediction loss is
\begin{equation}
\label{eqn:G_pred_u}  
    L_{G_{{\textrm{pred}_{(x_u,\hat{y}_u)}}}} = - \mathbb{E}_{x_u,\hat{y}_u \sim G}\log[1 - p(c_u = n+1 | (x_u,\hat{y}_u)].
\end{equation} 
Since the main objective of $G$ is to generate the segmentation map, a small weight is used for the adversarial loss terms for both labeled and unlabeled data. 

The class discriminator $D$ is trained on multiple objectives---adversary on the segmentation mask generator $G$'s output and classification of the images into the real or surrogate classes. Since the model is trained on both labeled and unlabeled training data, the loss function $L_D$ of the class discriminator $D$ includes both supervised and unsupervised losses. Function $L_D$ includes five different loss terms: 1) supervised classification loss on $(\hat{c}_l | x_l, y_l)$, 2) self-supervised classification loss on unlabeled data $(\hat{c}_u | x_u, \hat{y}_u)$, 3) adversarial real loss on $x_l, y_l$, 4) adversarial prediction loss on $x_l, \hat{y}_l$, and 5) adversarial prediction loss on $x_u, \hat{y}_u$. The supervised losses are calculated as
\begin{equation}
\label{eqn:D_sup_l}  
\begin{aligned}
    L_{D_{\textrm{sup}}} &= - \mathbb{E}_{x_l,y_l,c_l\sim p_{\mathcal{D_L}}} \\ 
    & \log\left[p(\hat{c}_l = (c_l = i) | x_l,y_l; i< n+1)\right]
    \end{aligned}
\end{equation}
and
\begin{equation}
\label{eqn:D_sup_u}  
\begin{aligned}
    L_{D_{\textrm{self}}} &= - \mathbb{E}_{x_u,\hat{y}_u,c_u\sim p_{\mathcal{D_U}}} \\ 
    & \log\left[p(\hat{c}_u = (c_u = i) | x_u,y_u; i< n+1)\right].
    \end{aligned}
\end{equation}
Now, for the labeled data $\mathcal{D}_L$, $D$ can receive two-way inputs $(x_l, y_l)$ or $(x_l, \hat{y}_l)$. Therefore, the adversarial losses for $\mathcal{D}_L$ are
\begin{equation}
\label{eqn:D_real_l}  
    L_{D_{\textrm{gt}}(\mathcal{D}_L)} = - \mathbb{E}_{x_l,y_l \sim p_{(\mathcal{D}_L)}} \log [ 1 - p(c_l = n+1 | x_l, y_l)]
\end{equation}
and
\begin{equation}
\label{eqn:D_pred_l}
L_{D_{\textrm{pred}}(\mathcal{D}_L)} = - \mathbb{E}_{(x_l,\hat{y}_l) \sim S} \log [p(c_l = n+1 | x_l,\hat{y}_l)].
\end{equation}
For the unlabeled data $\mathcal{D}_U$, we calculate only the adversarial prediction loss
\begin{equation}
\label{eqn:D_pred_u}
L_{D_{\textrm{pred}}(\mathcal{D}_U)} = - \mathbb{E}_{(x_u,\hat{y}_u) \sim S} \log [p(c_u = n+1 | x_u,\hat{y}_u)].
\end{equation}

\section{Experimental Evaluation}

\subsection{Datasets}

We evaluate our S${}^4$MTL model against the baseline models using the following two datasets:
\begin{enumerate}
    \item [1.] \textbf{Chest Dataset:} \hspace{2pt} For our experiments, we make use of the Montgomery County chest X-ray set, the Shenzhen chest X-ray set available from the NIH \cite{jaeger2014two}, and the dataset available from the Japanese Society of Radiological Technology (JCX) \cite{shiraishi2000development}. Combining these three publicly available datasets, we create a dataset of 912 chest X-ray images, which we call the ``Chest Dataset''. We split it into three sets---training (615), validation (69), and test (228). Along with lung segmentation, we perform 3-class classification---normal, tuberculosis (TB), and nodule.

\begin{figure}
\centering
 \resizebox{\linewidth}{!}{%
  \begin{tabular}{cccccc}
   \multirow{3}{*}{\includegraphics[width=\linewidth, trim={1cm 0cm 7cm 0cm},clip]{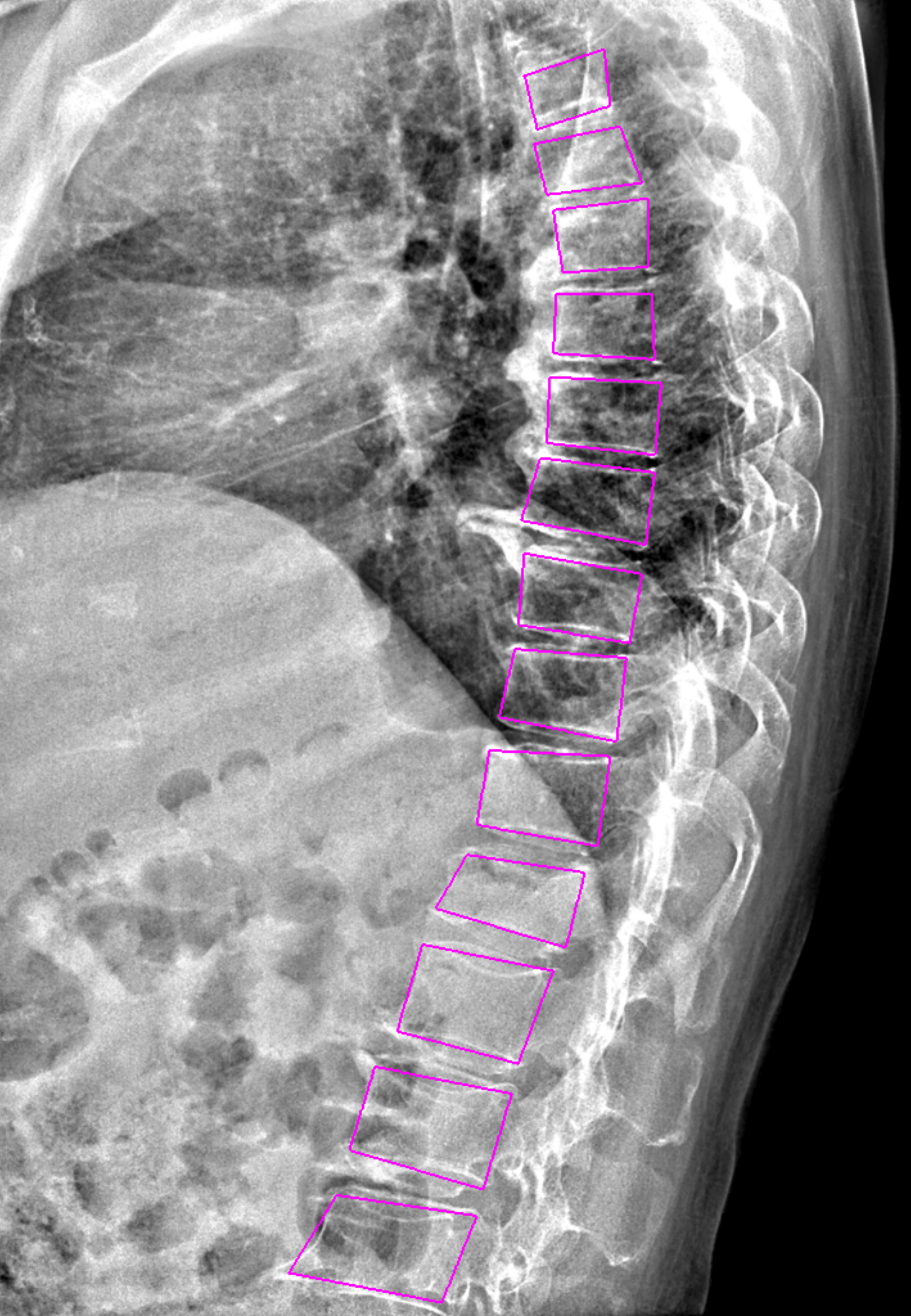}}
   \\
   \noalign{\smallskip}
    &
    \includegraphics[width=0.3\linewidth, trim={4cm 1cm 3cm 1cm},clip]{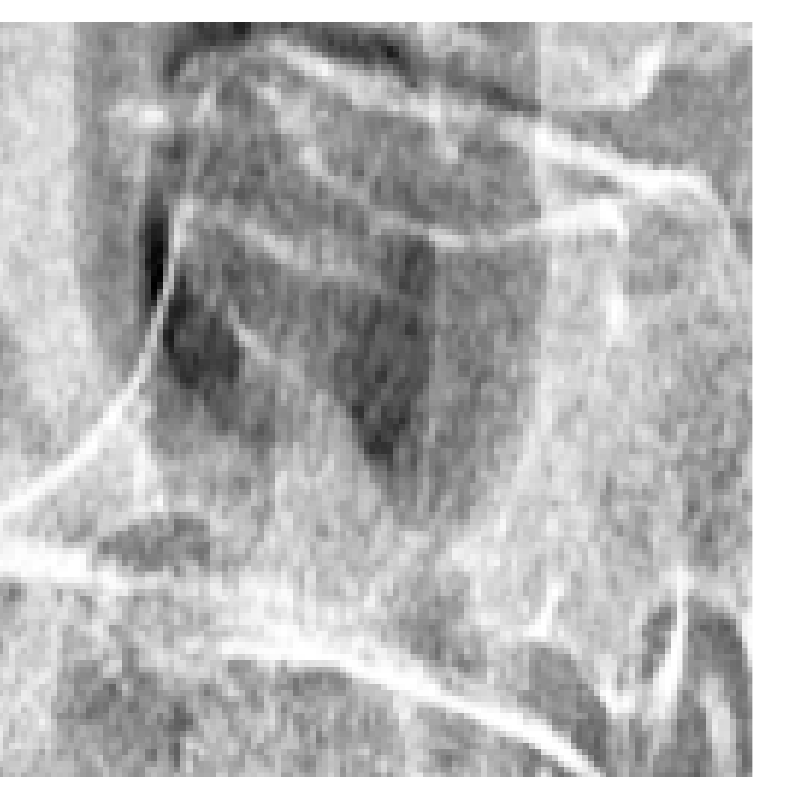}
    &
    \includegraphics[width=0.3\linewidth, trim={4cm 1cm 3cm 1cm},clip]{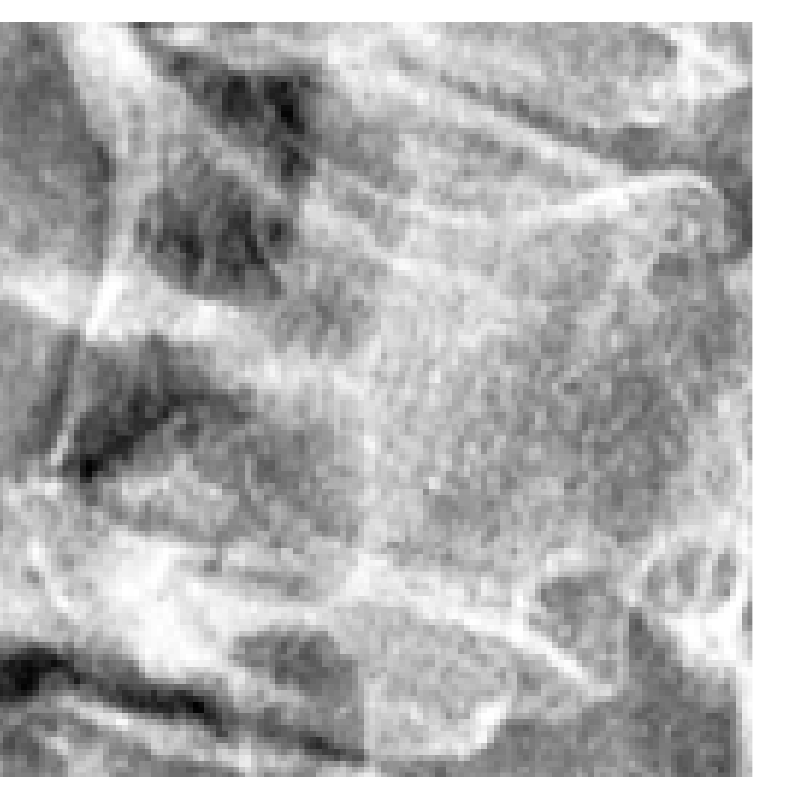}
    &
    \includegraphics[width=0.3\linewidth, trim={4cm 1cm 3cm 1cm},clip]{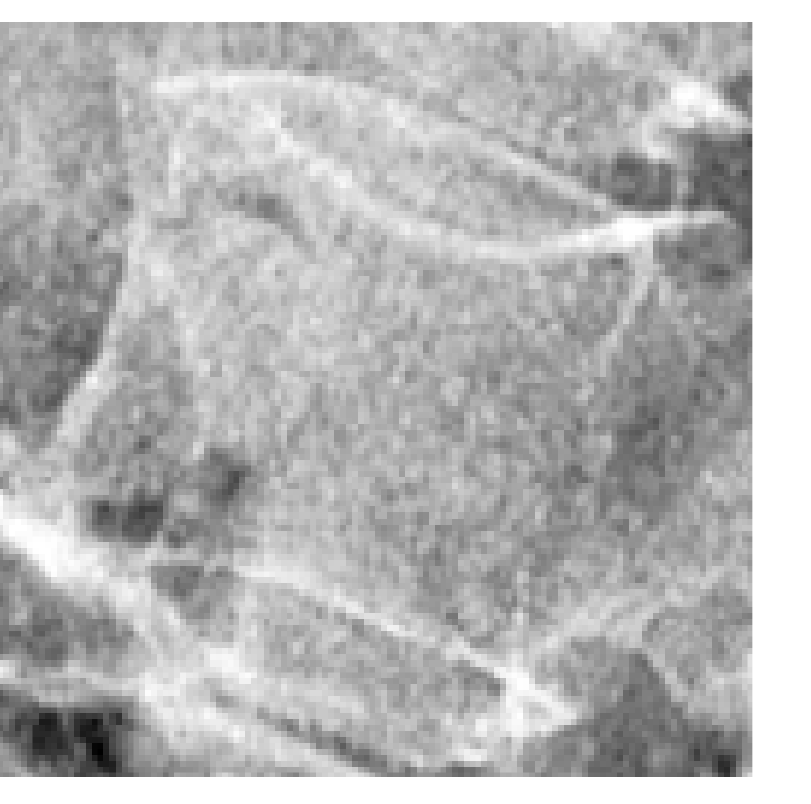}
    &
    \includegraphics[width=0.3\linewidth, trim={4cm 1cm 3cm 1cm},clip]{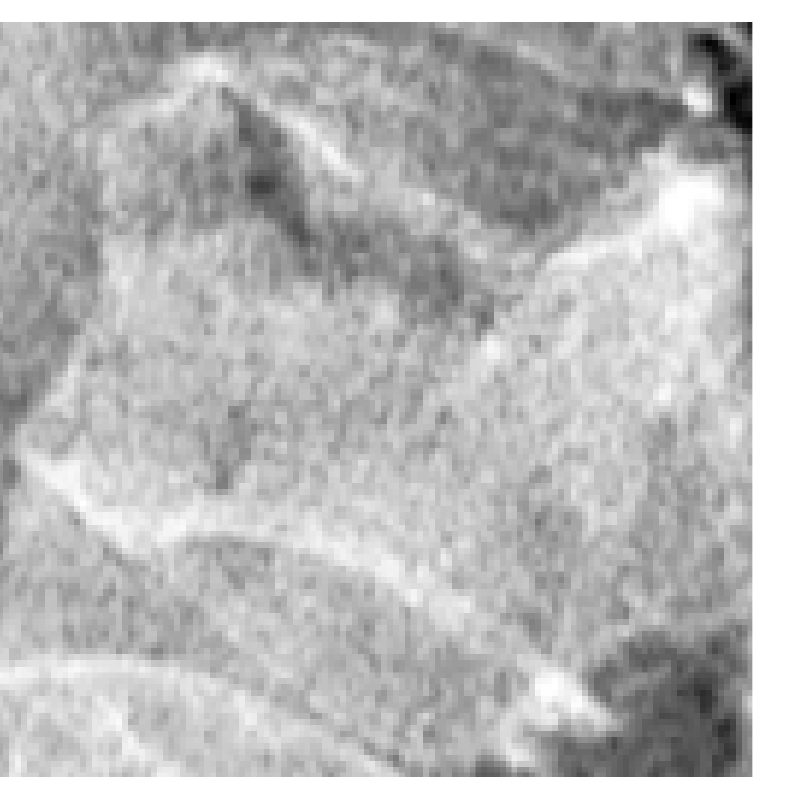}
    &
    \includegraphics[width=0.3\linewidth, trim={4cm 1cm 3cm 1cm},clip]{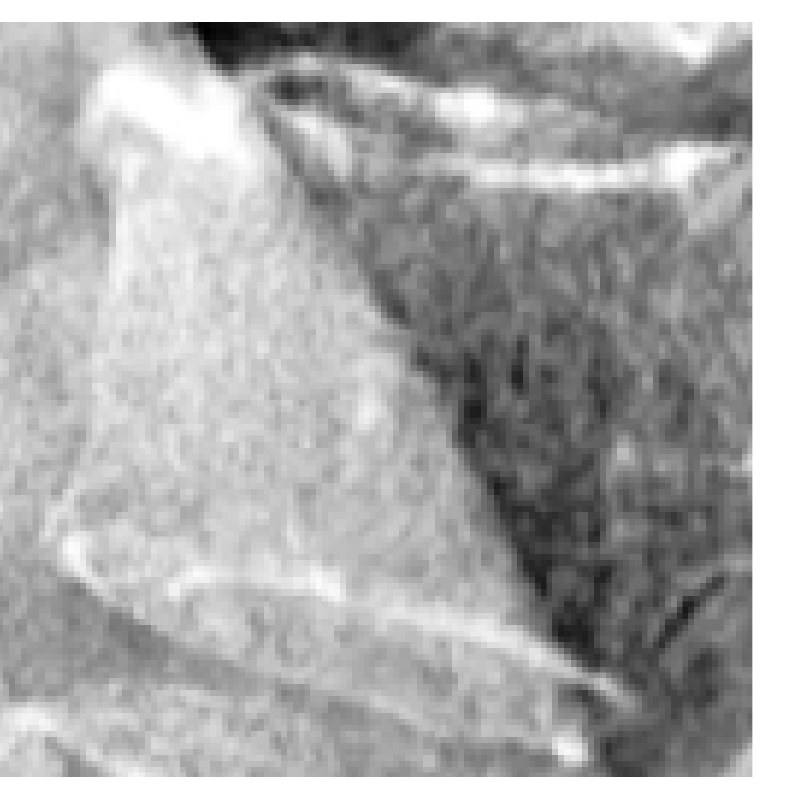}
    \\
    \noalign{\smallskip}
    &
    {\Huge L5} & {\Huge L4} & {\Huge L3} & {\Huge L2} & {\Huge L1} 
    \\
    \noalign{\smallskip}
    &
    \includegraphics[width=0.3\linewidth, trim={4cm 1cm 3cm 1cm},clip]{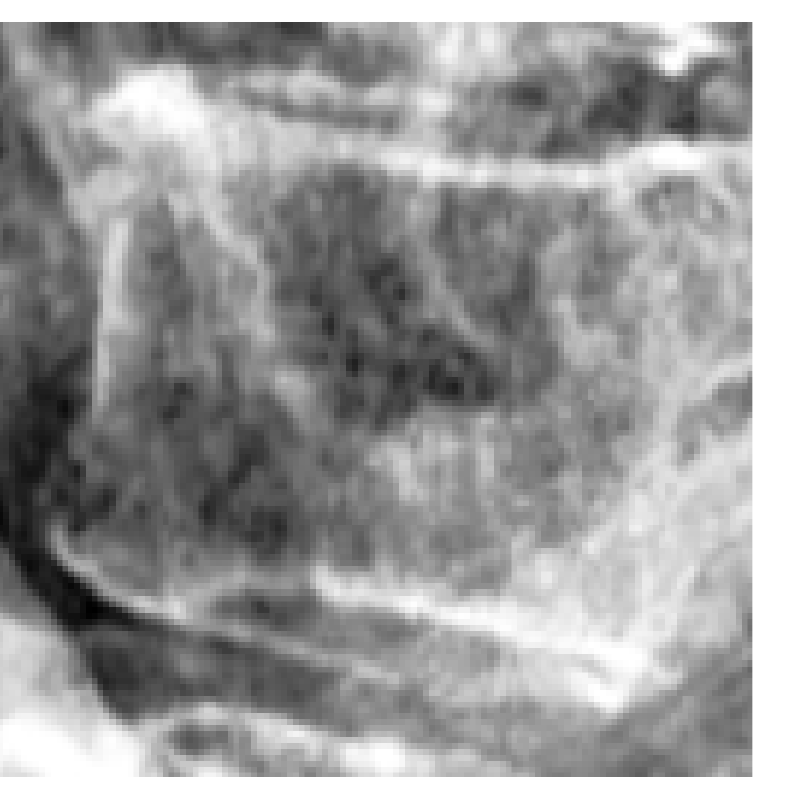}
    &
    \includegraphics[width=0.3\linewidth, trim={4cm 1cm 3cm 1cm},clip]{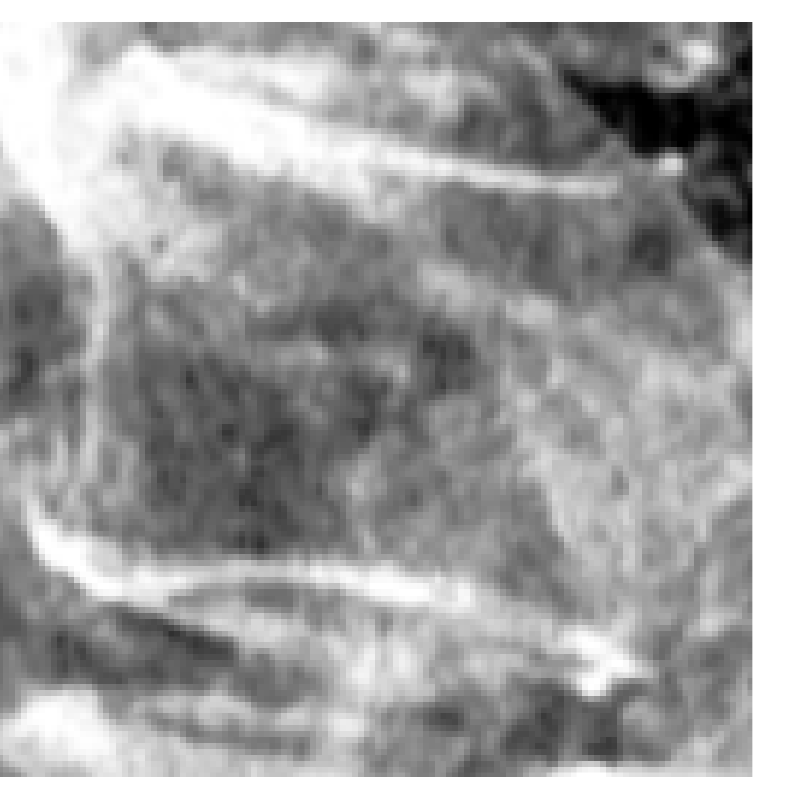}
    &
    \includegraphics[width=0.3\linewidth, trim={4cm 1cm 3cm 1cm},clip]{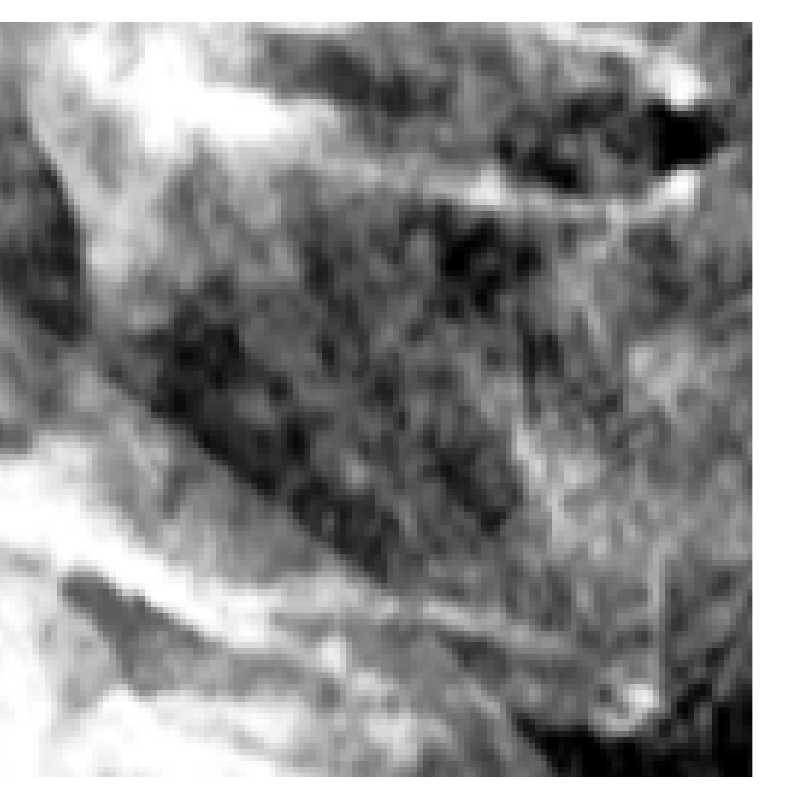}
    &
    \includegraphics[width=0.3\linewidth, trim={4cm 1cm 3cm 1cm},clip]{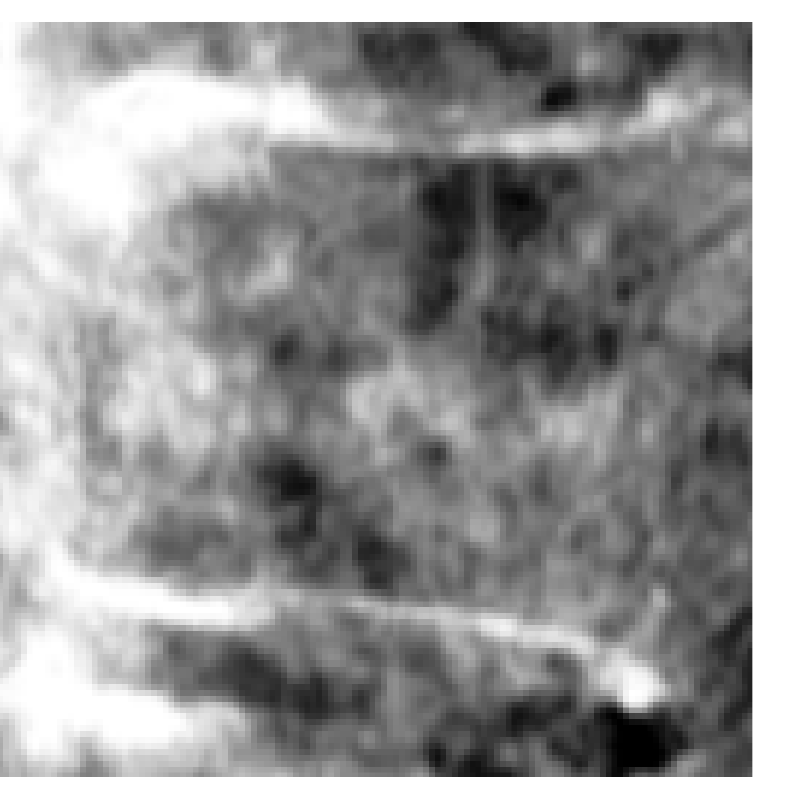}
    &
    \includegraphics[width=0.3\linewidth, trim={4cm 1cm 3cm 1cm},clip]{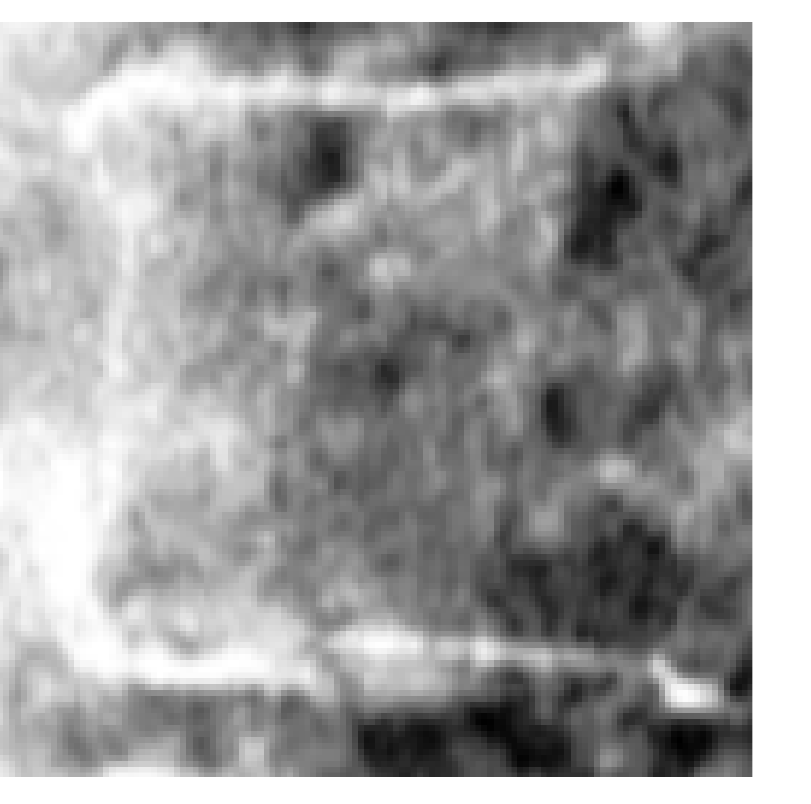}
    \\
    \noalign{\smallskip}
    &
    {\Huge T12} & {\Huge T11} & {\Huge T10} & {\Huge T9} & {\Huge T8}
    \\
    \noalign{\smallskip}
    &
    \includegraphics[width=0.3\linewidth, trim={4cm 1cm 3cm 1cm},clip]{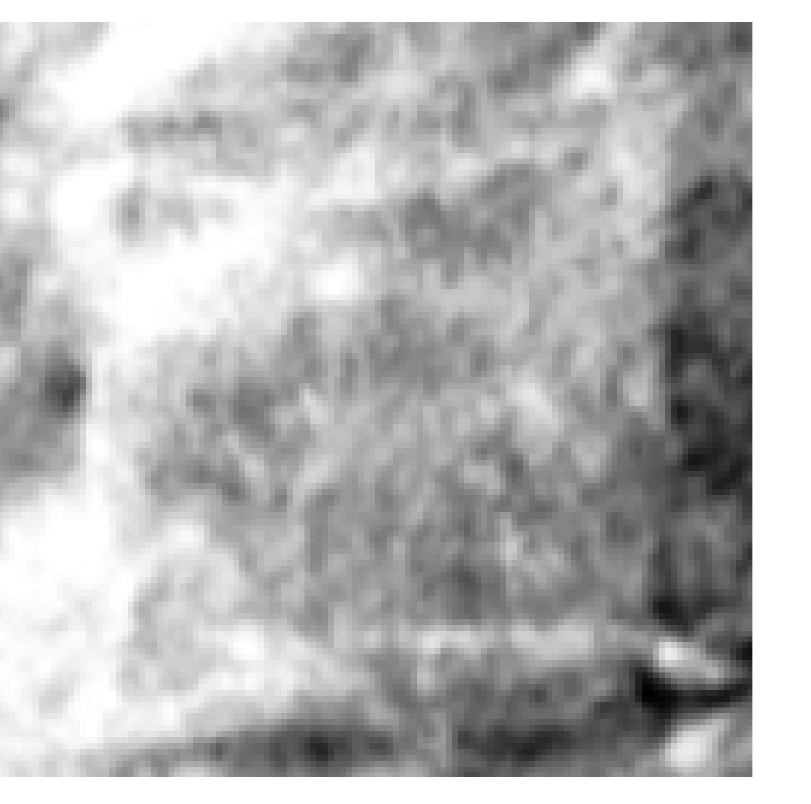}
    &
    \includegraphics[width=0.3\linewidth, trim={4cm 1cm 3cm 1cm},clip]{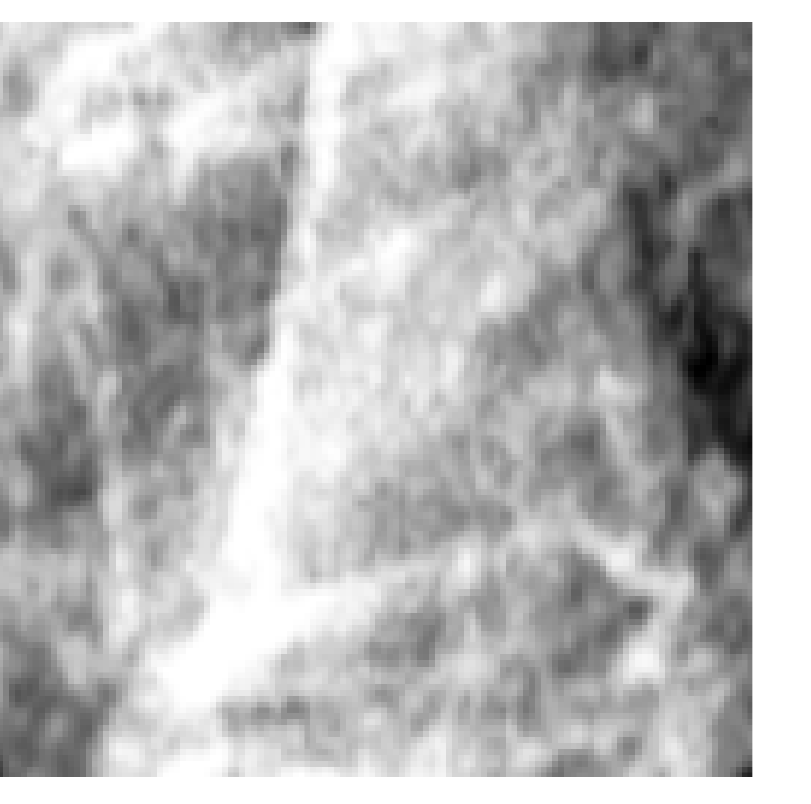}
    &
    \includegraphics[width=0.3\linewidth, trim={4cm 1cm 3cm 1cm},clip]{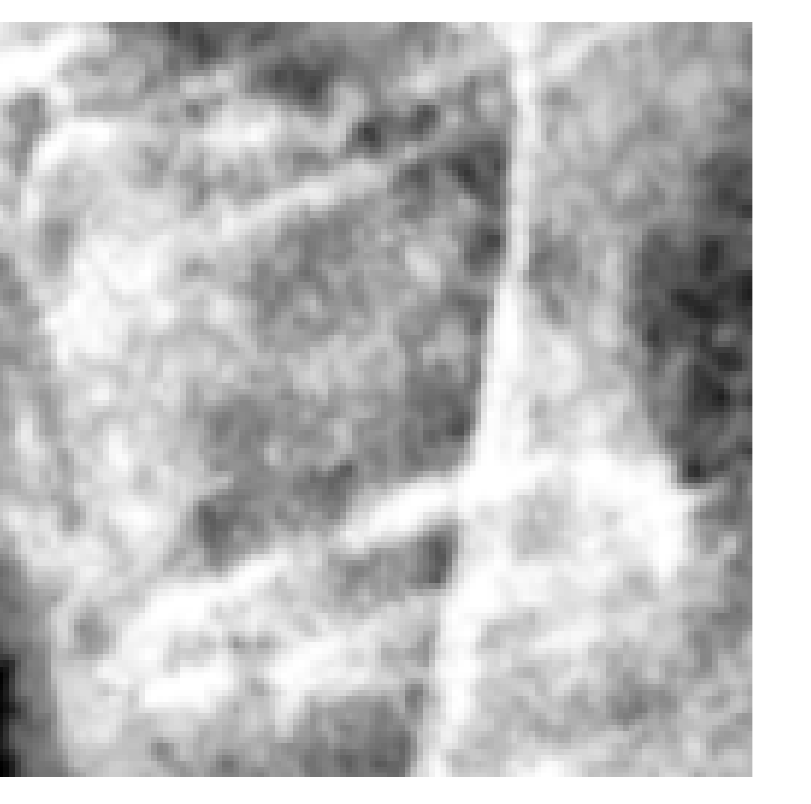}
    \\
    \noalign{\smallskip}
    &
    {\Huge T7} & {\Huge T6} & {\Huge T5}
    \\
    \end{tabular} }
    \smallskip
    \caption{Individual vertebra patches along with their labels (vertebra indices) extracted from the lateral X-ray image. L: Lumbar vertebra, T: Thoracic vertebra. L5 is the bottom most vertebra; other vertebrae are shown in reverse order. In this X-ray image, only L2 is labeled osteoporosis positive (abnormal) while the remaining vertebrae are normal.}
    \label{fig:lat-vert}
\end{figure}

    \item [2.] \textbf{Spine Dataset:} \hspace{2pt} A dataset of 100 very high-resolution spinal X-ray images is obtained from a vertebral compression fracture study of osteoporosis \cite{wong2013vertebral}. The dataset contains pixel-wise annotation and binary disease label (normal and abnormal) for each of the vertebrae in lateral views. We extract extended patches from each individual vertebra and create a dataset of 994 patch images, which we call the ``Spine Dataset''. Fig.~\ref{fig:lat-vert} demonstrates the extraction of vertebra patches from a spine X-ray image. The dataset was split into three subsets: train (713), validation (42), and test (139). We perform vertebra segmentation and abnormality prediction on each vertebra patch (normal vs abnormal).  
\end{enumerate}

For each dataset, we train the models on its training set (labeled and unlabeled data), use its validation set to determine the hyper-parameters and for model selection, and evaluate the models on its test set.

\subsection{Implementation Details}

\textbf{Inputs:} 
All the images are normalized and resized to $128\times128\times1$ before feeding them to the models. 

\textbf{Model Architecture:} As the segmentation mask generator we use a U-Net like encoder-decoder network with skip connections, and as the class discriminator we used another convolutional network (Conv-Net) \cite{imran2019semi}. We implement the S$^4$MTL algorithm in Tensorflow running on a Tesla P40 GPU and a 64-bit Intel(R) Xeon(R) 440G CPU. 

\textbf{Baselines:}  As baselines, we use the segmentation mask generator (U-Net) and class discriminator (Conv-Net) networks separately for single-task models both in supervised and partially-supervised manners. Using the same backbone network, we also train a multitasking U-Net with a classification branch from its bottleneck layer (similarly as Y-Net \cite{Mehta2018YNetJS}), namely U-MTL. In addition, we experiment with another semi-supervised multitasking model without self-supervision and unsupervised segmentation losses. We call this the semi-supervised multitask learning (S$^2$MTL) model.  

\textbf{Training:} Following our formulated problem, $\mathcal{D_L}$ and $\mathcal{D_U}$ are selected before training the models, rather than just masking some data out during training. We constrain all the semi-supervised models to having maximum 50\% labeled data in order to hold $|\mathcal{D_L}| \leq |\mathcal{D_U}|$. The semi-supervised models (single-task or multitask) are trained on varying proportions of labeled data: 5\%, 10\%, 20\%, 30\%, and 50\%. For example, when 10\% is selected, 10\% of the training data are used with their corresponding class and segmentation labels, and 90\% are used without any label information. We adapt the training signal annealing in our model using the logarithmic schedule \cite{xie2019unsupervised} with an adjustment for balancing between epochs and mini-batches. In our experiments, the signal annealing threshold $\eta_{(e,s)}$ is set to $1-\exp{(-\frac{s*e+1}{E*N})}*(1-\frac{1}{n}) + \frac{1}{n}$, where $s$ is the current step, $e$ is the current epoch, $E$ is total number of epochs, $N$ is the training dataset size, and $n$ is the number of classes. In training, the labeled data are selected such that every class has equal representation in the training data $\mathcal{D_L}$. In the mask generator network, we use instance-normalization and ReLU activation. A dropout rate of 0.4 is applied after every convolutional layer.

\begin{table*}
\setlength{\tabcolsep}{4pt}
\centering
\caption{Classification performance comparison of the S${}^4$MTL model against the baseline models in different data settings with varying proportions of labeled data.}
\medskip
\label{table:class}
\resizebox{0.7\linewidth}{!}{
\begin{tabular}{c l | c c c ? c c c c}
            \toprule
           \multirow{2}{*}{\hspace{8pt}Type}
            &
           \multirow{2}{*}{\hspace{8pt}Model}
           &
           \multicolumn{3}{c?}{Spine}
           &
           \multicolumn{4}{c}{Chest}
           \\
          &&
           Accuracy
           &
           F1(Normal)
           &
           F1(Abnormal)
           &
           Accuracy
           &
           F1(Normal)
           & 
           F1(TB)
           &
           F1(Nodule)
           \\
           \midrule
           \multirow{6}{*}{\rotatebox{45}{Single-Task}}
           &
           Conv-Net-100\% 
           &
           0.780&0.740&0.800&{\bf0.700}&0.530&0.720&0.840 
           \\
           &
           Conv-Net-50\% 
           &
           0.620&0.680&0.580&0.630&0.460&0.710&0.730 
           \\
           &
           Conv-Net-30\% 
           &
           0.500&0.000 &0.630&0.580&0.590&0.360&0.710
           \\
           &
           Conv-Net-20\% 
           &
           0.460& 0.460&0.460 &0.570&0.620&0.250&0.730
           \\
           &
           Conv-Net-10\% 
           &
           0.450& 0.490&0.410&0.560&0.630&0.180&0.710 
           \\
           &
           Conv-Net-5\% 
           &
           0.420 & 0.330&0.490&0.480&0.650&0.000&0.000 
           \\
           \midrule
           \multirow{16}{*}{\rotatebox{45}{Multitask}}
           &
           UMTL-100\%
           & 
           0.683 & 0.810 & 0.008&0.650&0.610&0.290&0.780  
           \\
           &
           UMTL-50\%
           & 
           0.520 & 0.580 & 0.430&0.500&0.680&0.200&0.000  
           \\
           &
           UMTL-30\%
           & 
           0.670 & 0.800 & 0.000&0.480&0.650&0.000&0.000  
           \\
           &
           UMTL-20\%
           & 
           0.660 & 0.800 & 0.000&0.450&0.530&0.450&0.000  
           \\
           &
           UMTL-10\%
           & 
           0.350 & 0.030 & 0.051 & 0.482 & 0.650 & 0.000 & 0.000  
           \\
           &
           UMTL-5\%
           & 
           0.331 & 0.000 & 0.050&0.480&0.650&0.000&0.000  
           \\
           \cmidrule{2-9}
           &
           S$^2$MTL-50\%
           & 
           0.670 & 0.800 & 0.000&0.650&0.610&0.290&0.780  
           \\
           &
           S$^2$MTL-30\%
           & 
           0.550 & 0.680 & 0.250&0.580&0.590&0.380&0.690  
           \\
           &
           S$^2$MTL-20\%
           & 
           0.500 & 0.000 & 0.630&0.500&0.570&0.410&0.490  
           \\
           &
           S$^2$MTL-10\%
           & 
           0.460 & 0.430 & 0.490&0.480&0.650&0.000&0.000  
           \\
           &
          S$^2$MTL-5\%
           & 
           0.420 & 0.330 & 0.500&0.390&0.470&0.350&0.280  
           \\
           \cmidrule{2-9}
           &
          S$^4$MTL-50\%
           & 
           {\bf0.800} & 0.790 & 0.810&{\bf0.700}&0.530&0.710&0.730  
           \\
           &
           S$^4$MTL-30\%
           & 
           0.740 & 0.810 & 0.700&0.650&0.780&0.310&0.400  
           \\
           &
           S$^4$MTL-20\%
           & 
           0.680 & 0.790 & 0.290 &0.630&0.630&0.290&0.410 
           \\
           &
           S$^4$MTL-10\%
           & 
           0.670 & 0.800 & 0.000&0.620&0.620&0.300&0.840  
           \\
           &
          S$^4$MTL-5\%
           & 
           0.630 & 0.730 & 0.410&0.590&0.600&0.360&0.710  
           \\
           \bottomrule
        \end{tabular}
}
\end{table*}

\textbf{Hyper-parameters:} We use the Adam optimizer with adaptive learning rates for $G$ and $D$. With initial learning rates $2e-3$ with momentum 0.9 for $G$ and $1e-4$ with momentum 0.6 for $D$. The learning rates are adapted with exponential decay scheduled after every 2 epoch with decay rates of 0.9 and 0.5, respectively. Each model is trained with a mini-batch size of 16. 

\textbf{Evaluation:} For classification, along with the overall accuracy, we record the class-wise F1 scores. Dice similarity (DS), average Hausdorff distance (HD), Jaccard index (JI), structural similarity measure (SSIM), precision (Prec), and recall (Rec) scores are used to evaluate the segmentation performance.

\subsection{Results \& Discussion}

\paragraph{Classification:} We validate our S${}^4$MTL model on two separate datasets and the performance evaluations are compared against semi-supervised and fully-supervised models. The consistent improvement of S${}4$MTL with varying labeled data proportions is evident in Fig.~\ref{fig:class_plots}. For both the Spine and Chest datasets, our S$^4$MTL model is found to be superior to all the baseline models. Table~\ref{table:class} compares the classification performances among all the models, and our model achieves better overall and class-wise accuracies for both datasets, even better than the fully-supervised single-task and multi-task models.

\begin{figure}
\centering
 \resizebox{\linewidth}{!}{%
  \begin{tabular}{ccccccc}
   {\Large \rotatebox{90}{GT}} & \includegraphics[width=0.2\linewidth, trim={4cm 1cm 3cm 1cm}, clip]{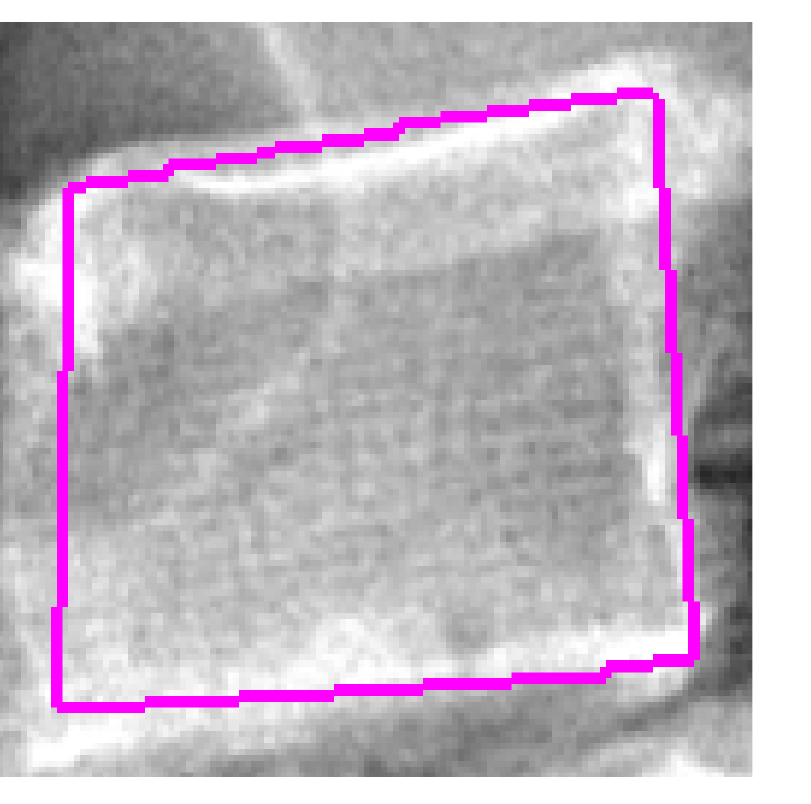}
   \smallskip
   \\
    &
    {\Large 5\%} & {\Large 10\%} & {\Large 20\%} & {\Large 30\%} & {\Large 50\%} & {\Large 100\%}\\
    {\Large \rotatebox{90}{U-Net}} 
    &
    \includegraphics[width=0.2\linewidth, trim={4cm 1cm 3cm 1cm},clip]{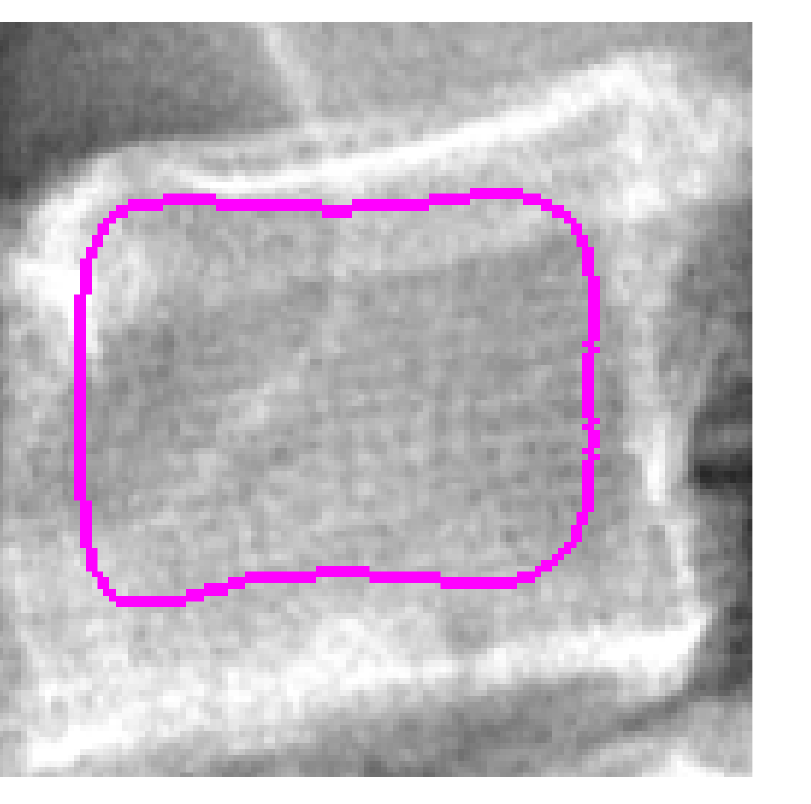}
    &
    \includegraphics[width=0.2\linewidth, trim={4cm 1cm 3cm 1cm},clip]{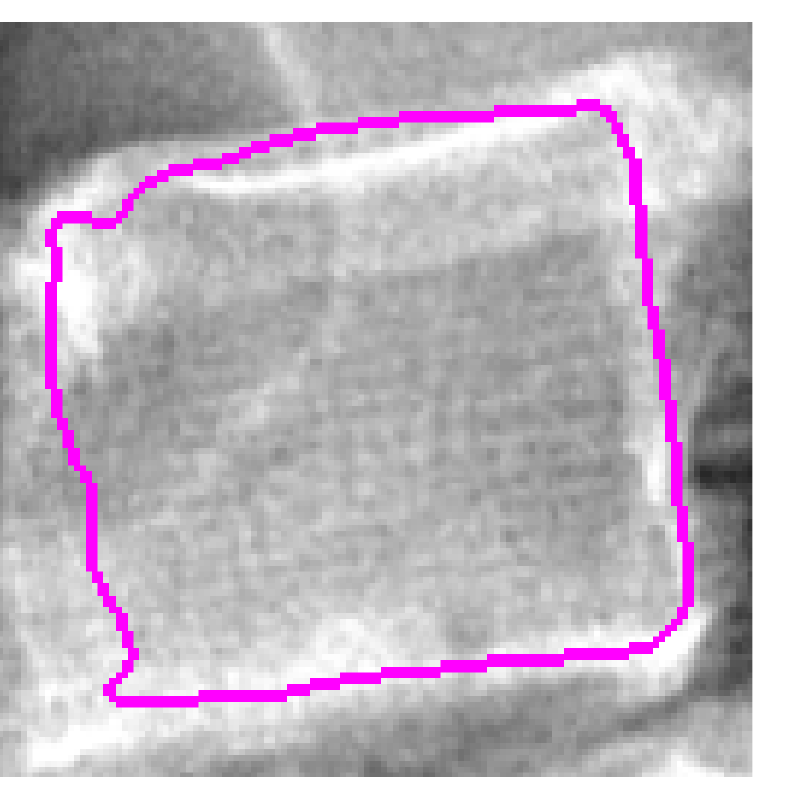}
    &
    \includegraphics[width=0.2\linewidth, trim={4cm 1cm 3cm 1cm},clip]{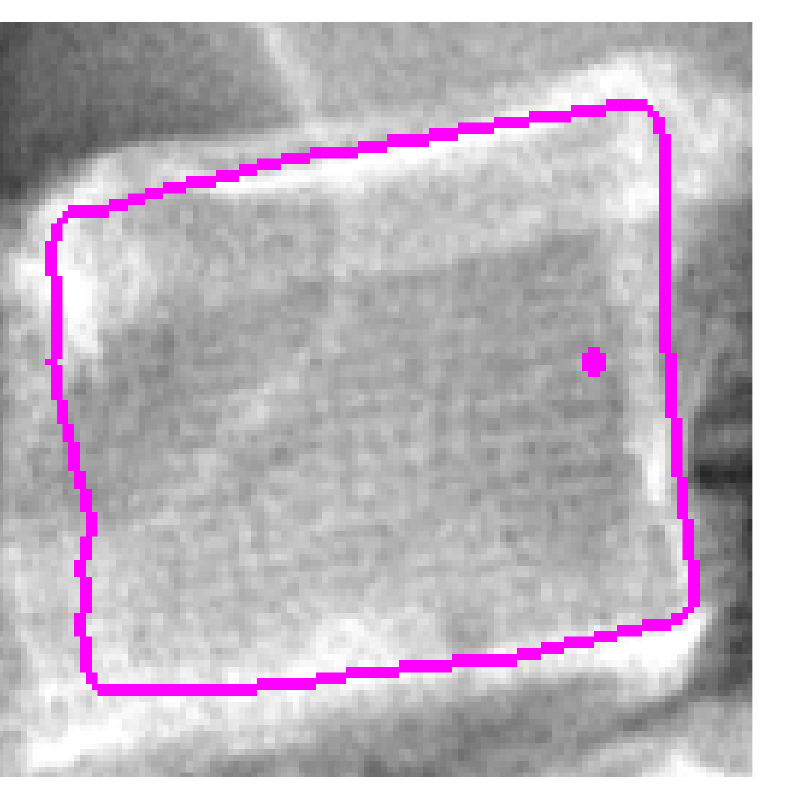}
    &
    \includegraphics[width=0.2\linewidth, trim={4cm 1cm 3cm 1cm},clip]{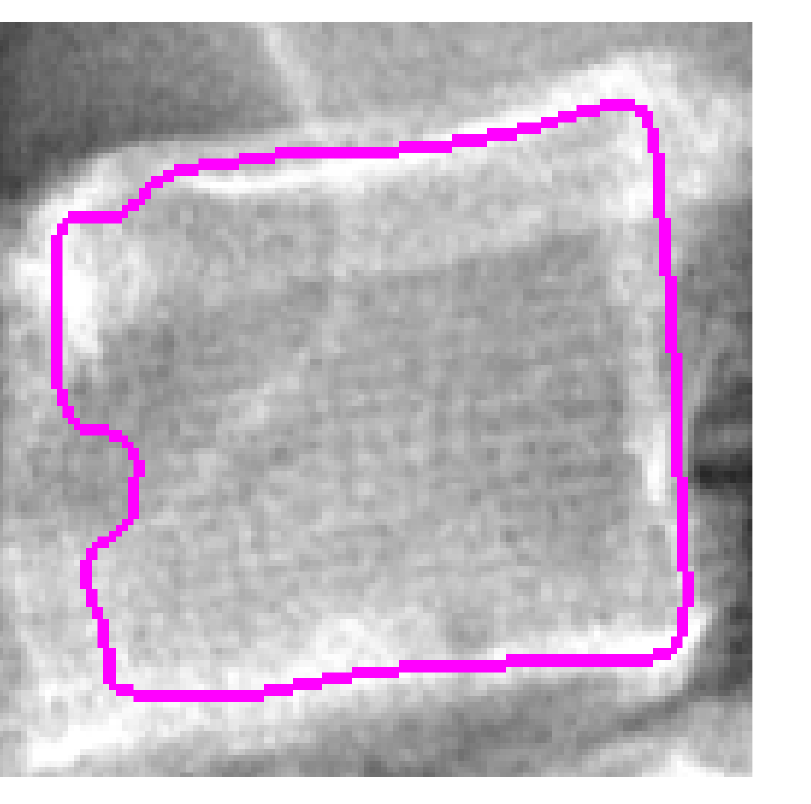}
    &
    \includegraphics[width=0.2\linewidth, trim={4cm 1cm 3cm 1cm},clip]{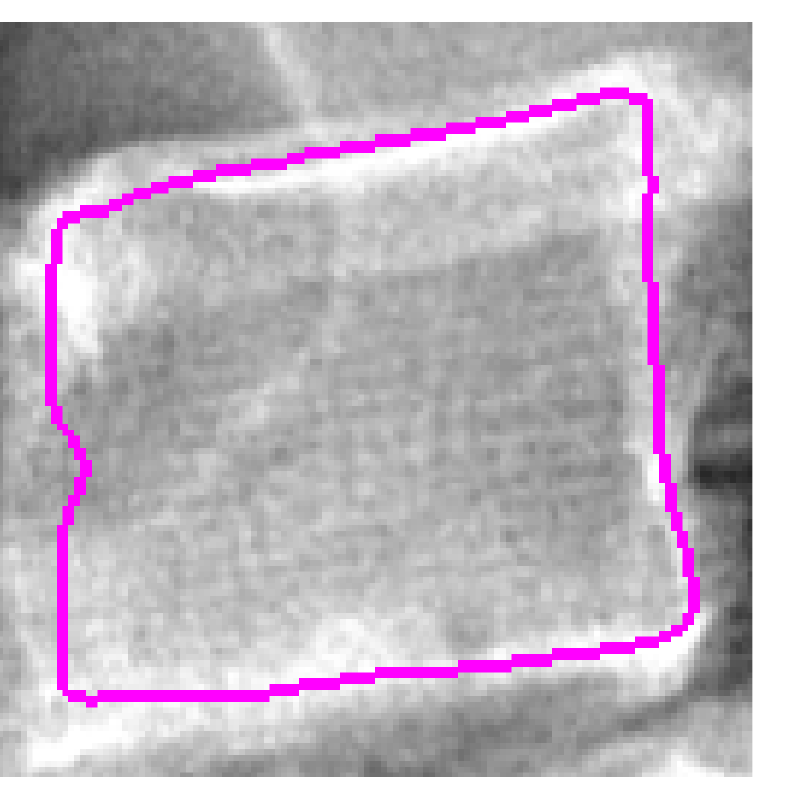}
    &
    \includegraphics[width=0.2\linewidth, trim={4cm 1cm 3cm 1cm},clip]{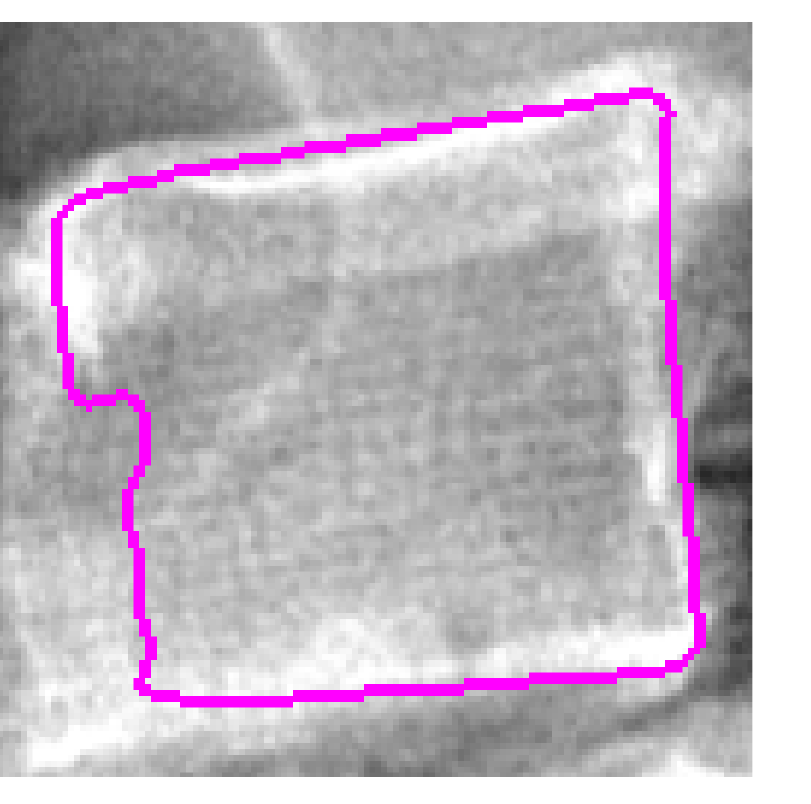}
    \\
    \noalign{\smallskip}
    {\Large \rotatebox{90}{U-MTL}} 
    &
    \includegraphics[width=0.2\linewidth, trim={4cm 1cm 3cm 1cm},clip]{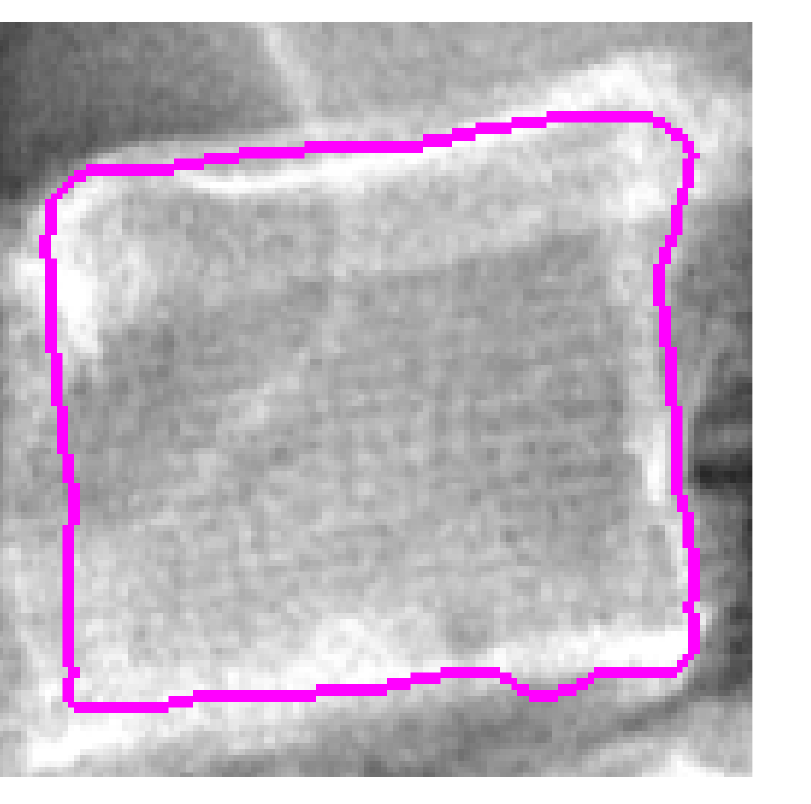}
    &
    \includegraphics[width=0.2\linewidth, trim={4cm 1cm 3cm 1cm},clip]{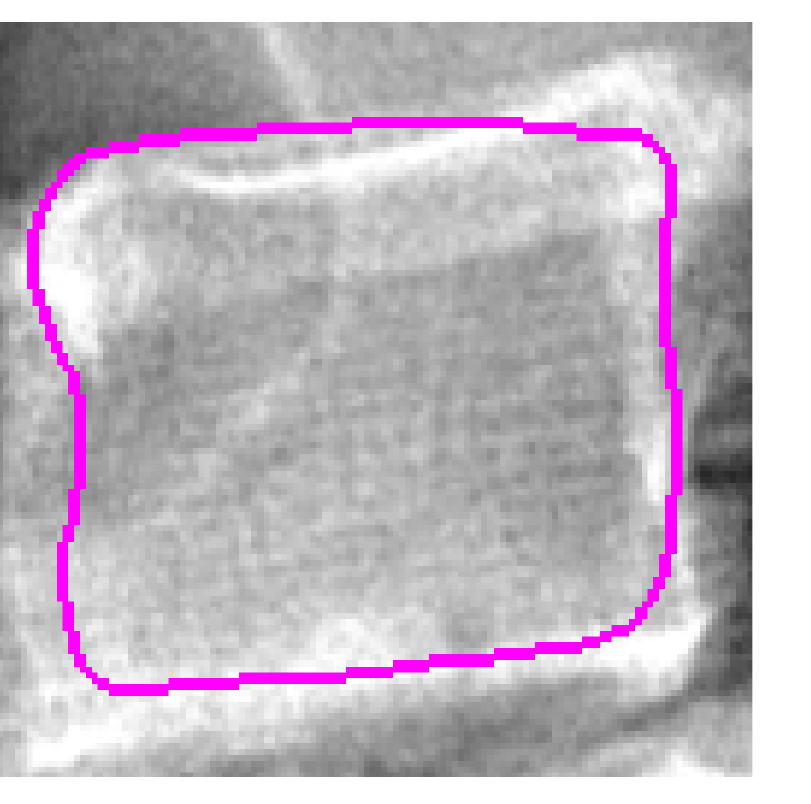}
    &
    \includegraphics[width=0.2\linewidth, trim={4cm 1cm 3cm 1cm},clip]{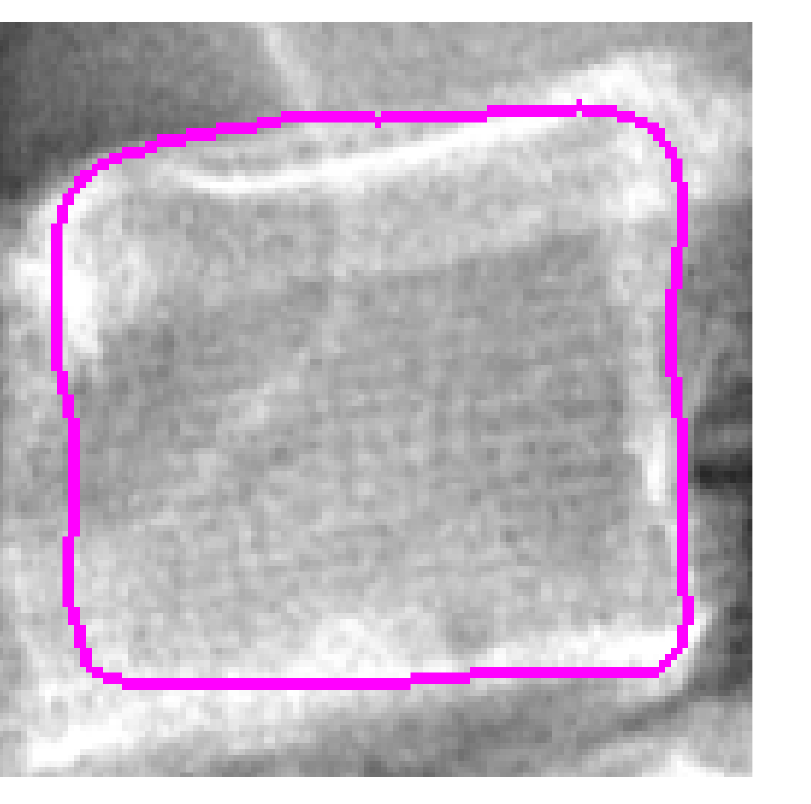}
    &
    \includegraphics[width=0.2\linewidth, trim={4cm 1cm 3cm 1cm},clip]{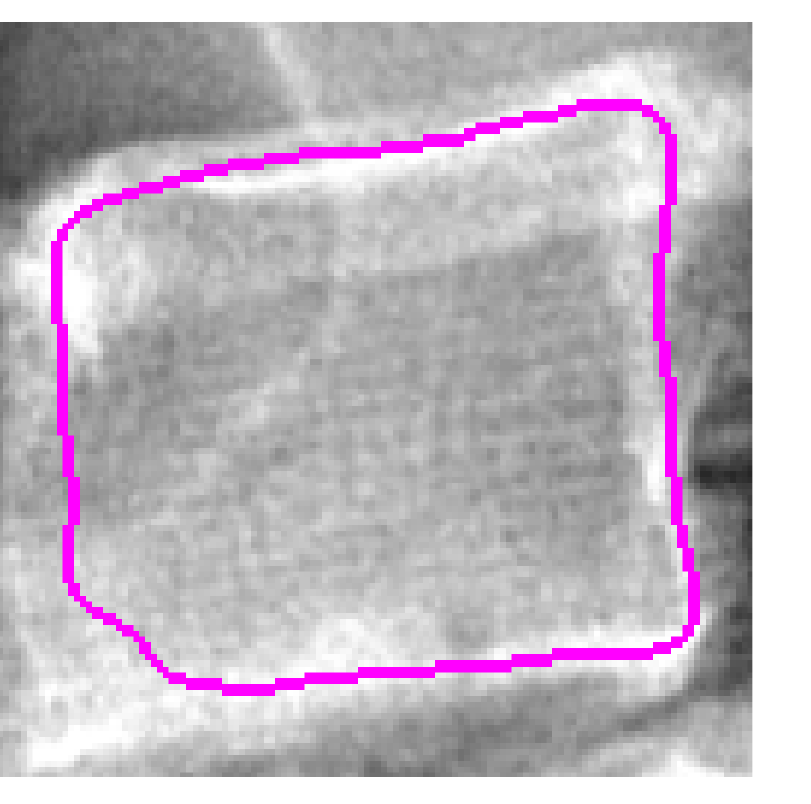}
    &
    \includegraphics[width=0.2\linewidth, trim={4cm 1cm 3cm 1cm},clip]{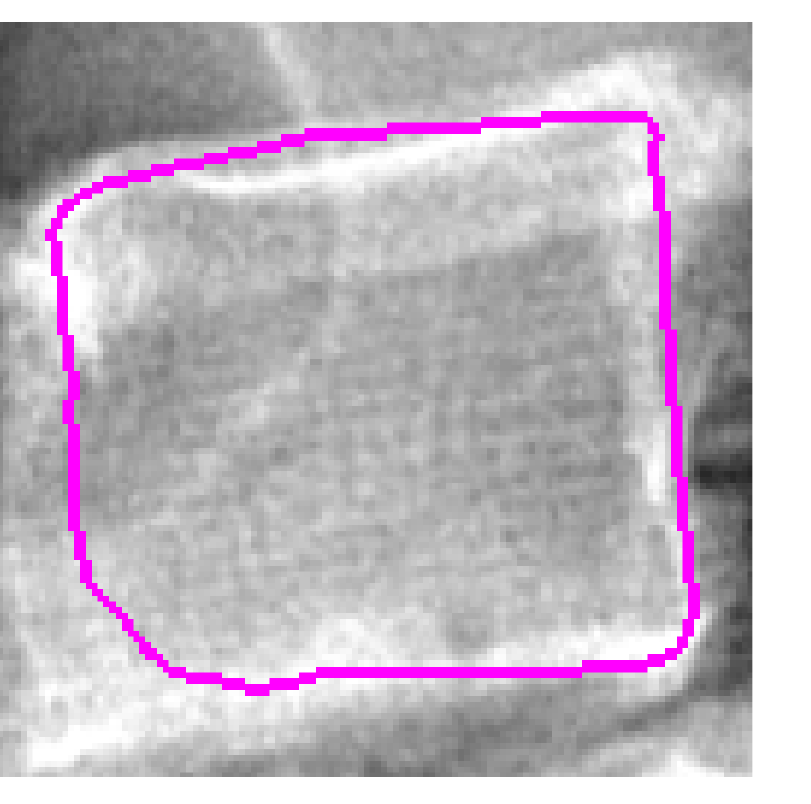}
    &
    \includegraphics[width=0.2\linewidth, trim={4cm 1cm 3cm 1cm},clip]{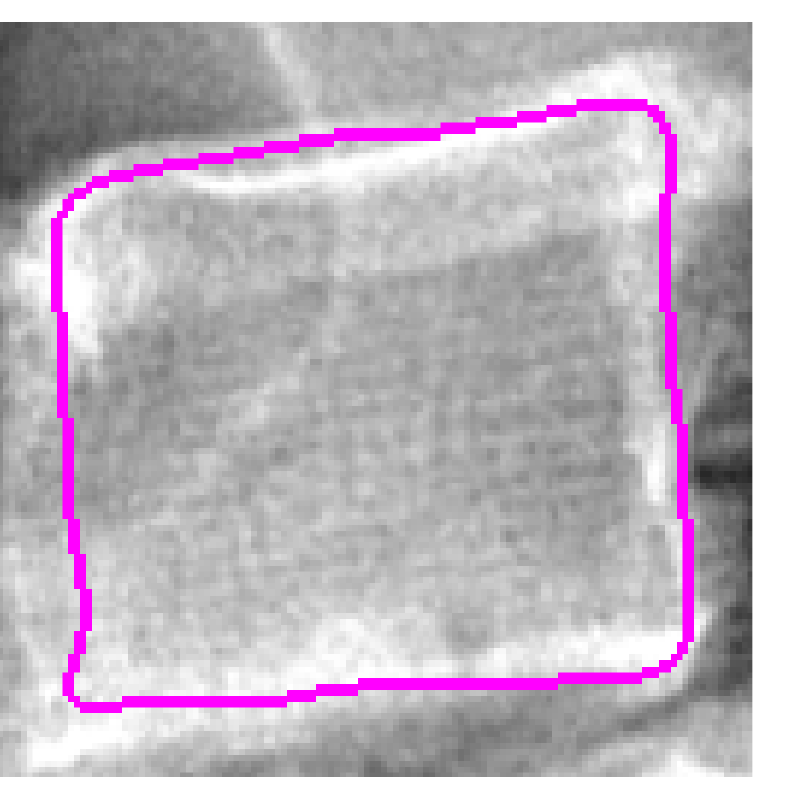}
    \\
    \noalign{\smallskip}
    {\Large \rotatebox{90}{S${}^2$MTL}} 
    &
    \includegraphics[width=0.2\linewidth, trim={4cm 1cm 3cm 1cm},clip]{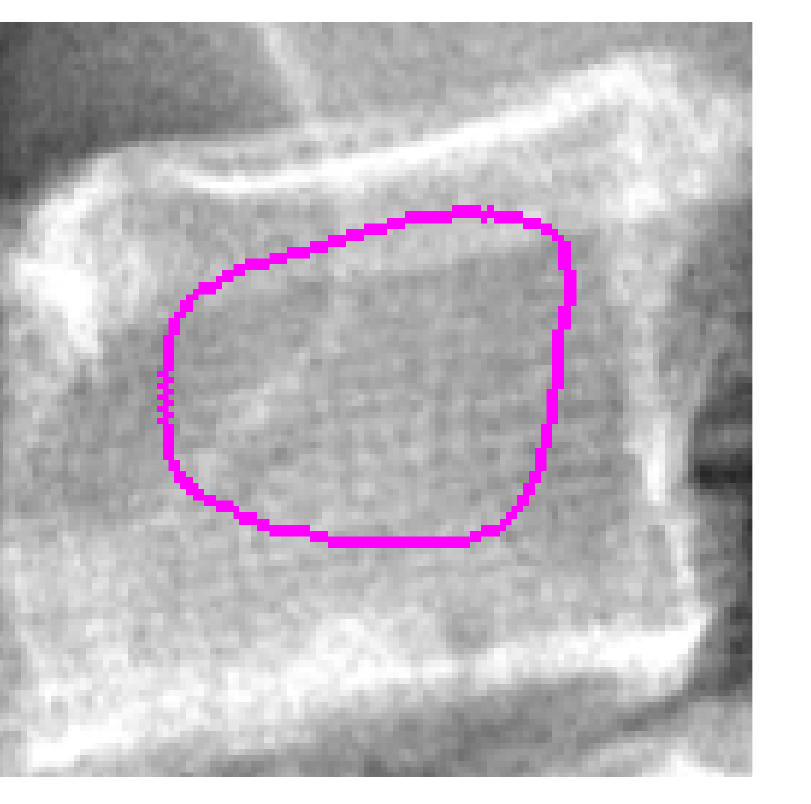}
    &
    \includegraphics[width=0.2\linewidth, trim={4cm 1cm 3cm 1cm},clip]{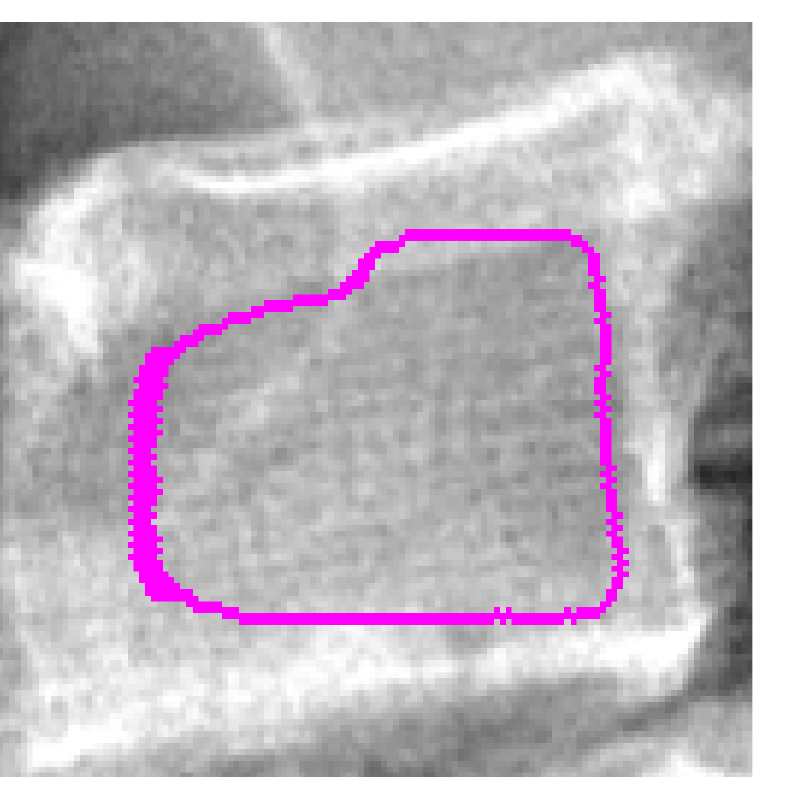}
    &
    \includegraphics[width=0.2\linewidth, trim={4cm 1cm 3cm 1cm},clip]{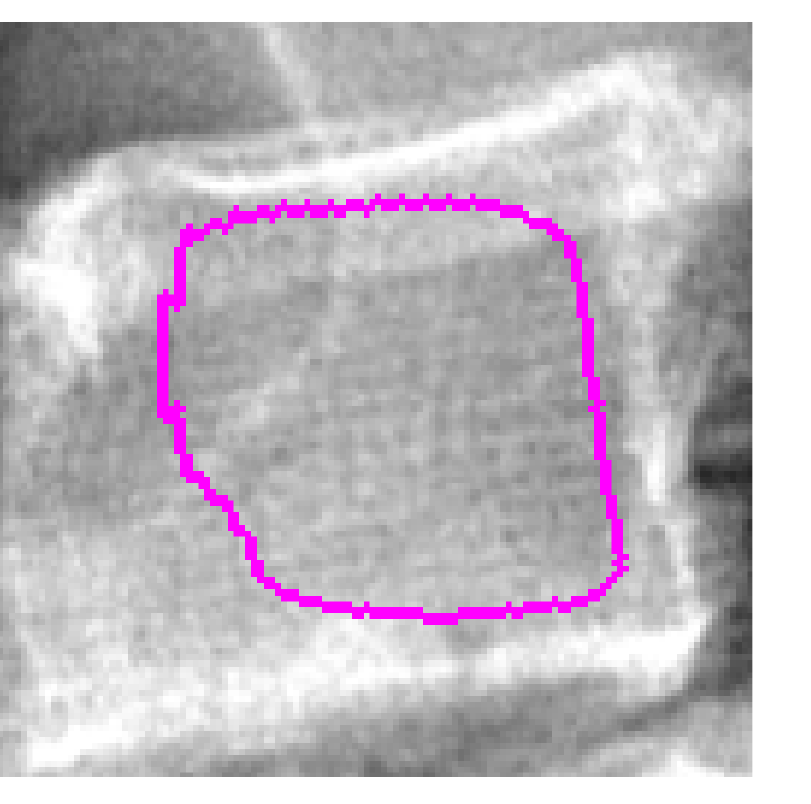}
    &
    \includegraphics[width=0.2\linewidth, trim={4cm 1cm 3cm 1cm},clip]{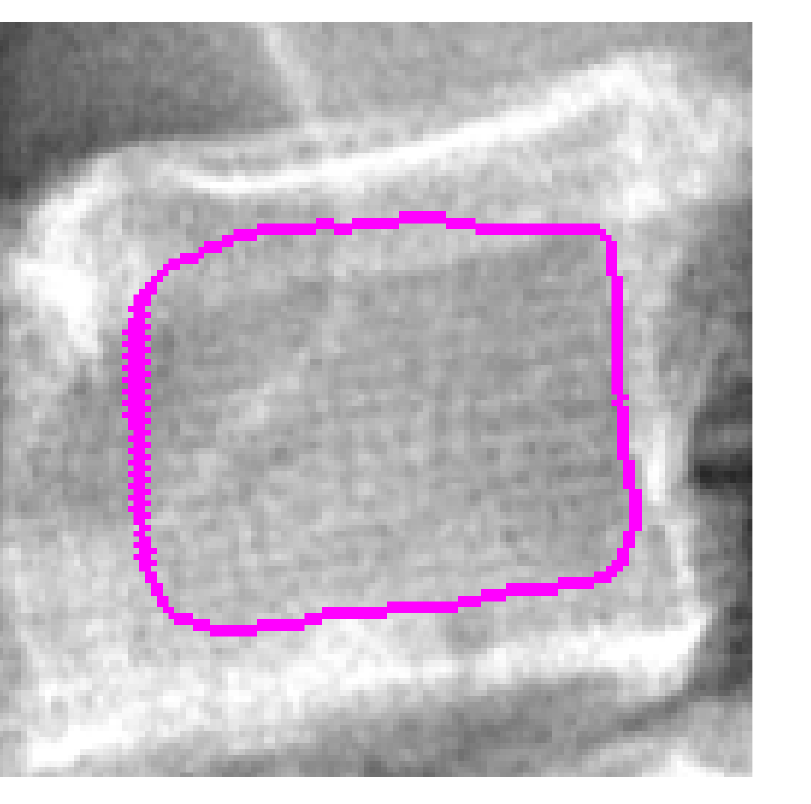}
    &
    \includegraphics[width=0.2\linewidth, trim={4cm 1cm 3cm 1cm},clip]{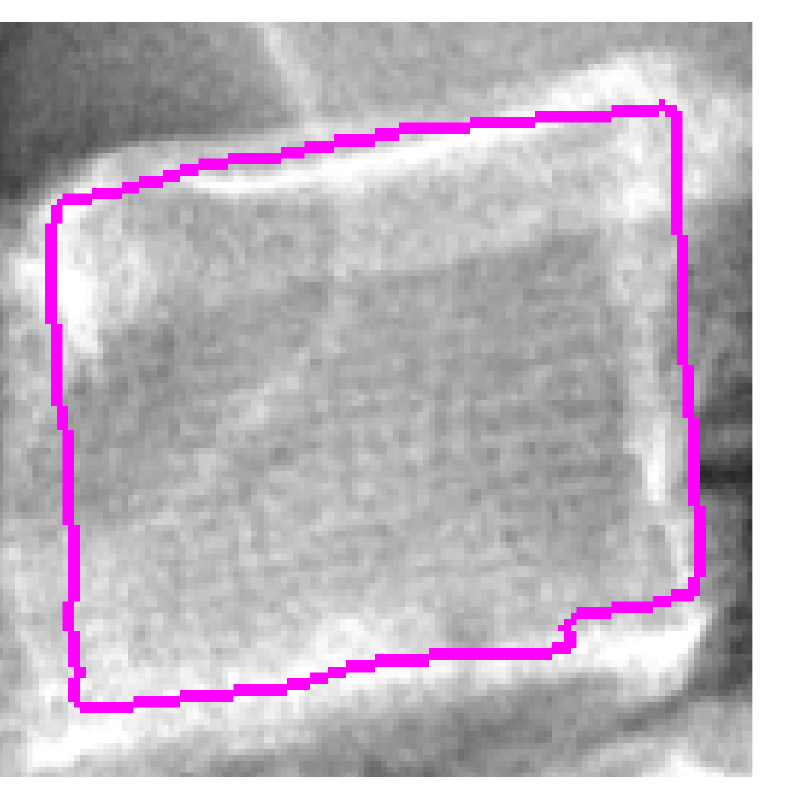}
    \\
    \noalign{\smallskip}
    {\Large \rotatebox{90}{S${}^4$MTL}} 
    &
    \includegraphics[width=0.2\linewidth, trim={4cm 1cm 3cm 1cm},clip]{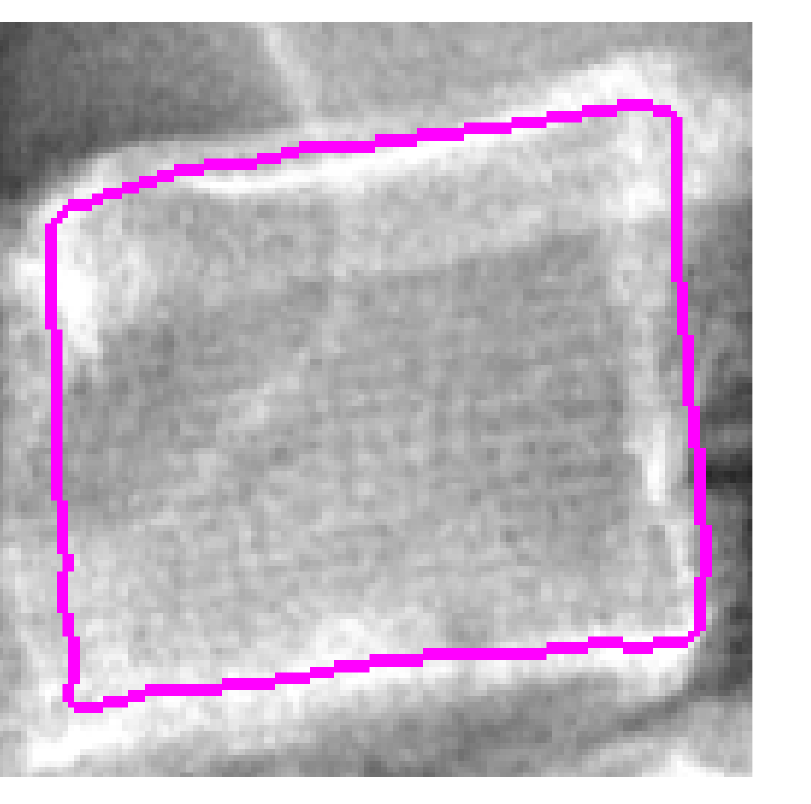}
    &
    \includegraphics[width=0.2\linewidth, trim={4cm 1cm 3cm 1cm},clip]{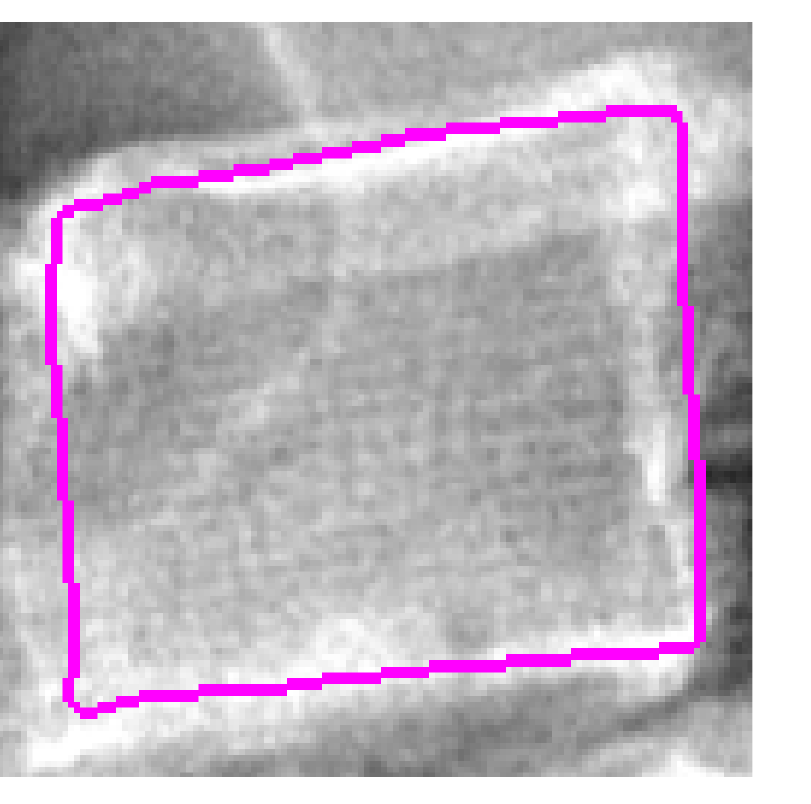}
    &
    \includegraphics[width=0.2\linewidth, trim={4cm 1cm 3cm 1cm},clip]{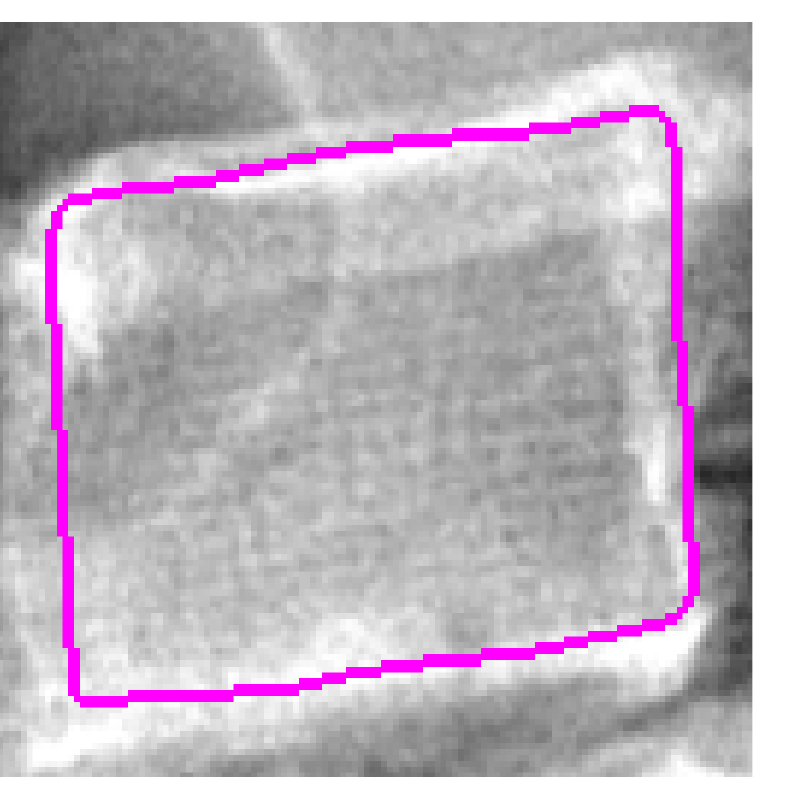}
    &
    \includegraphics[width=0.2\linewidth, trim={4cm 1cm 3cm 1cm},clip]{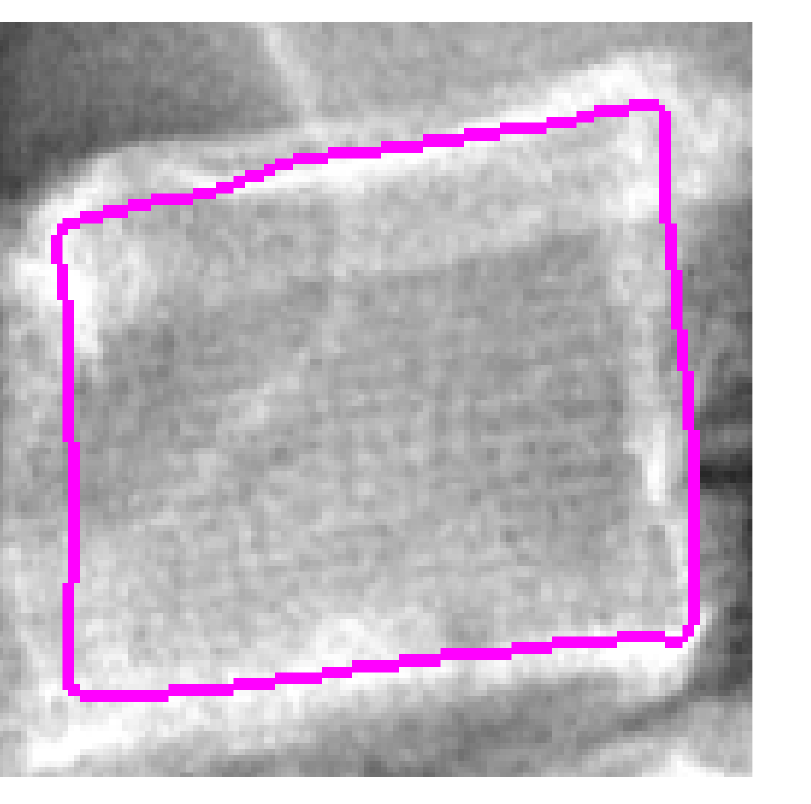}
    &
    \includegraphics[width=0.2\linewidth, trim={4cm 1cm 3cm 1cm},clip]{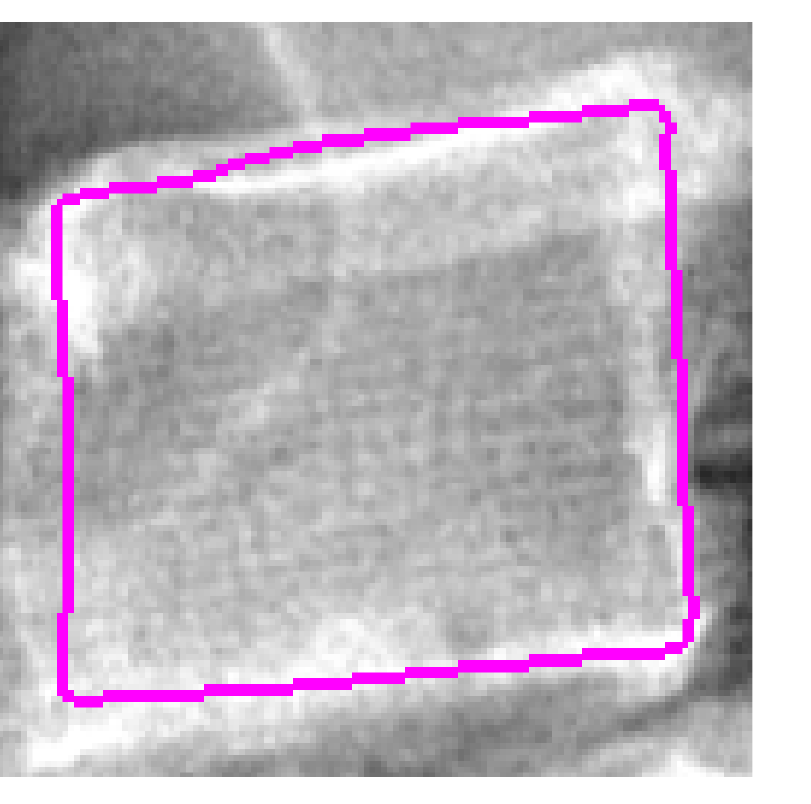}
    \\
    \end{tabular} }
    \caption{Spine Dataset: Boundary visualization of a predicted vertebra mask showing the superiority of our S$^4$MTL model over the baseline models with varying proportions of labeled data.}
    \label{fig:spine_vis}
\end{figure}

\begin{figure}
\centering
 \resizebox{\linewidth}{!}{%
  \begin{tabular}{ccccccc}
    {\Large \rotatebox{90}{GT}} & \includegraphics[width=0.2\linewidth, trim={4cm 1cm 3cm 1cm},clip]{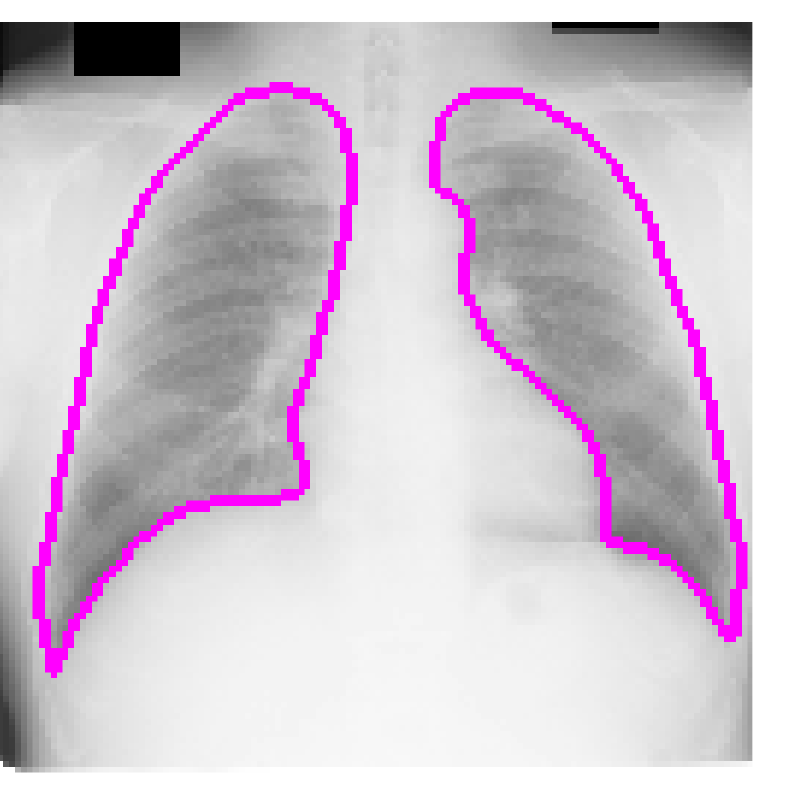}
    \smallskip
    \\
    &
    {\Large 5\%} & {\Large 10\%} & {\Large 20\%} & {\Large 30\%} & {\Large 50\%} & {\Large 100\%}\\
    {\Large \rotatebox{90}{U-Net}} 
    &
    \includegraphics[width=0.2\linewidth, trim={4cm 1cm 3cm 1cm},clip]{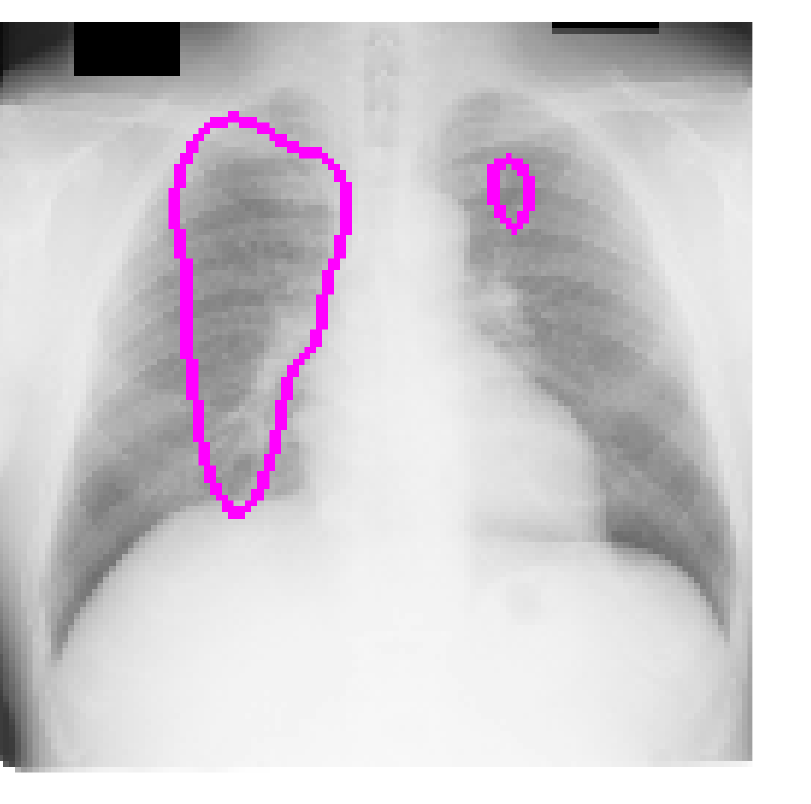}
    &
    \includegraphics[width=0.2\linewidth, trim={4cm 1cm 3cm 1cm},clip]{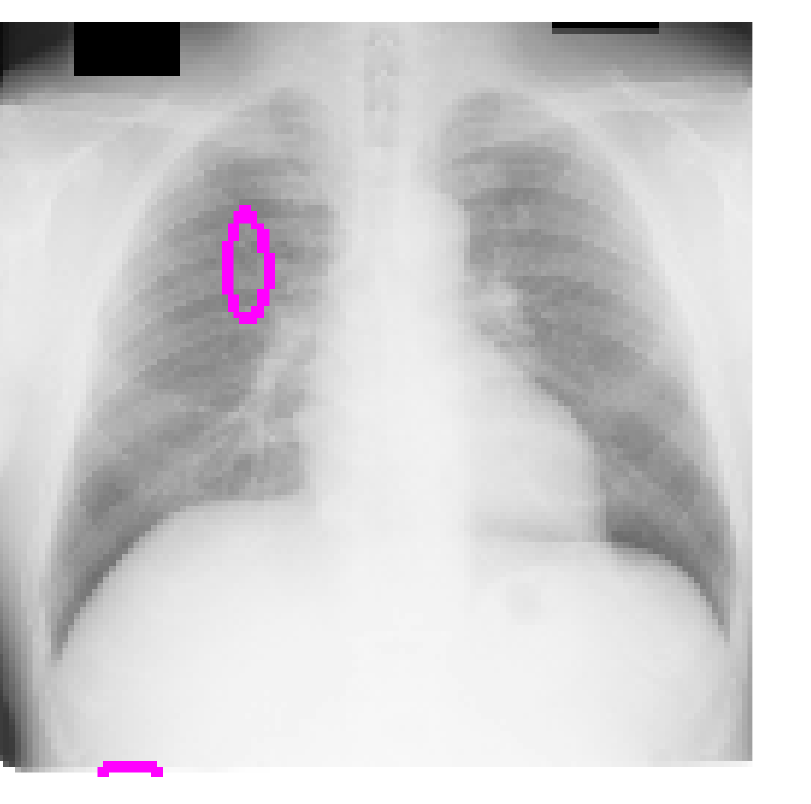}
    &
    \includegraphics[width=0.2\linewidth, trim={4cm 1cm 3cm 1cm},clip]{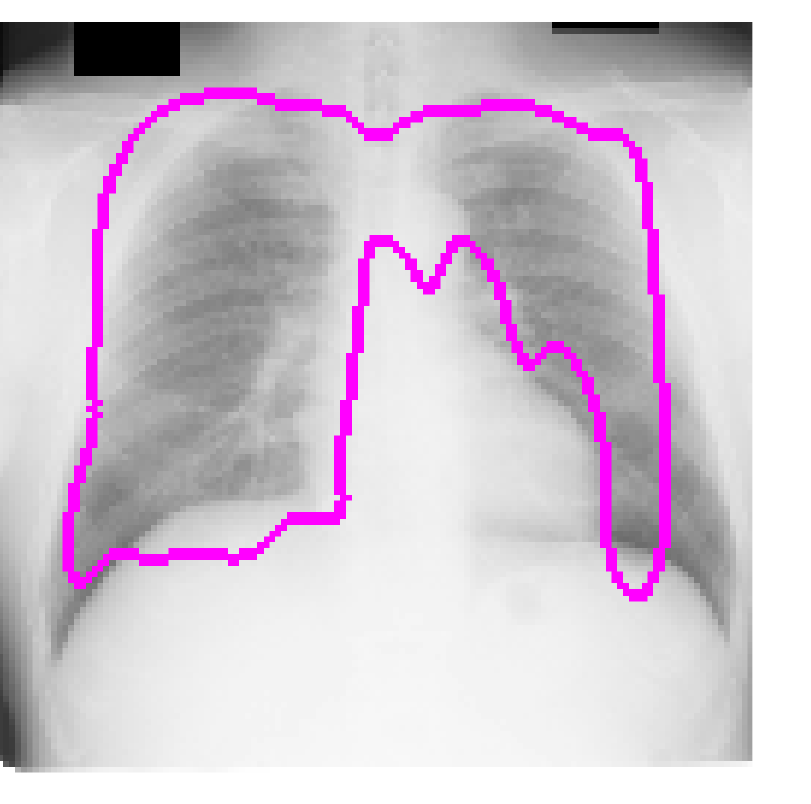}
    &
    \includegraphics[width=0.2\linewidth, trim={4cm 1cm 3cm 1cm},clip]{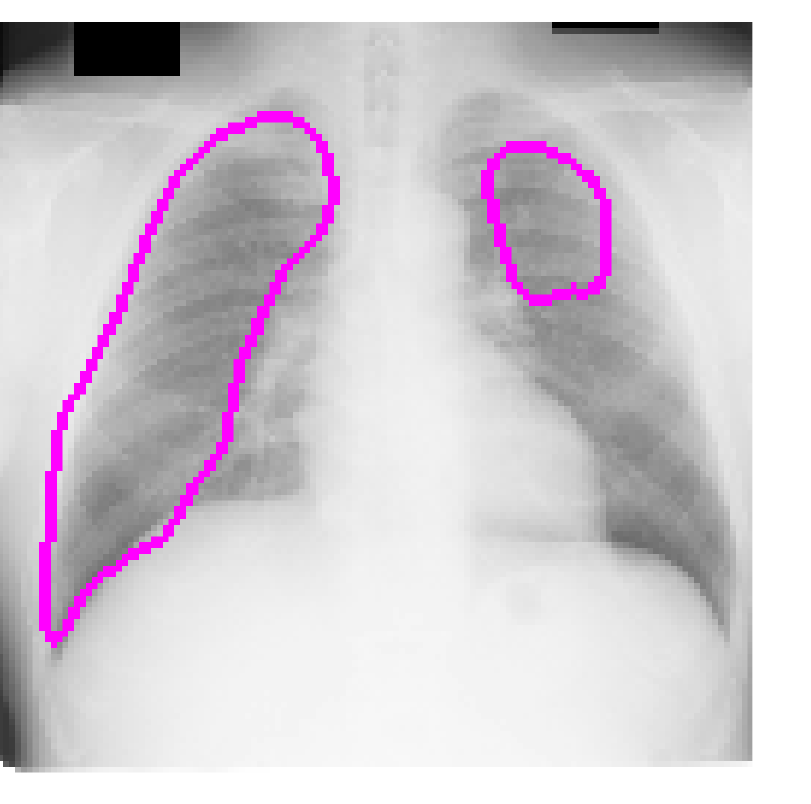}
    &
    \includegraphics[width=0.2\linewidth, trim={4cm 1cm 3cm 1cm},clip]{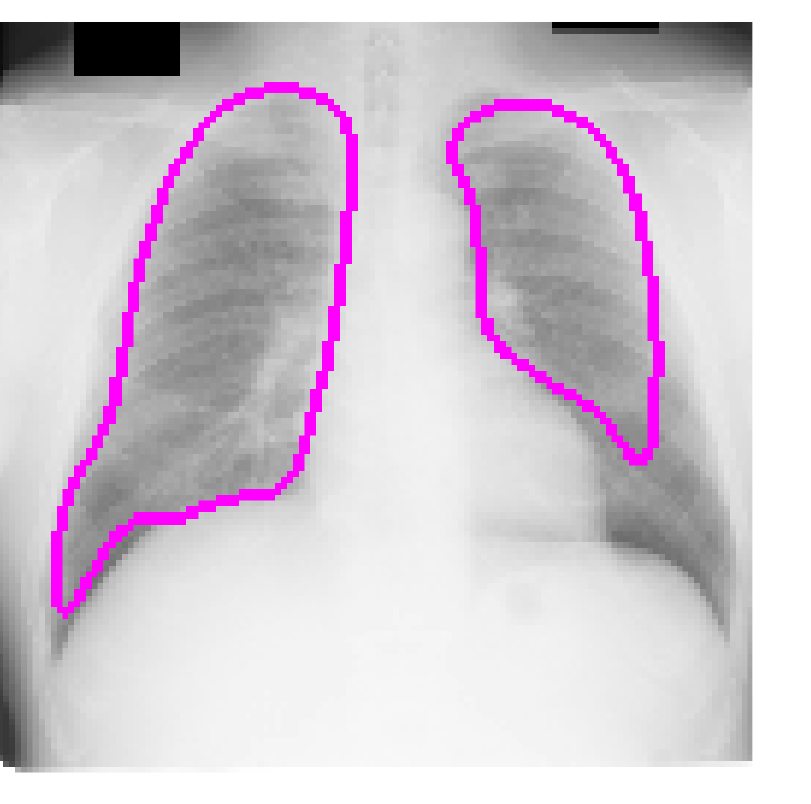}
    &
    \includegraphics[width=0.2\linewidth, trim={4cm 1cm 3cm 1cm},clip]{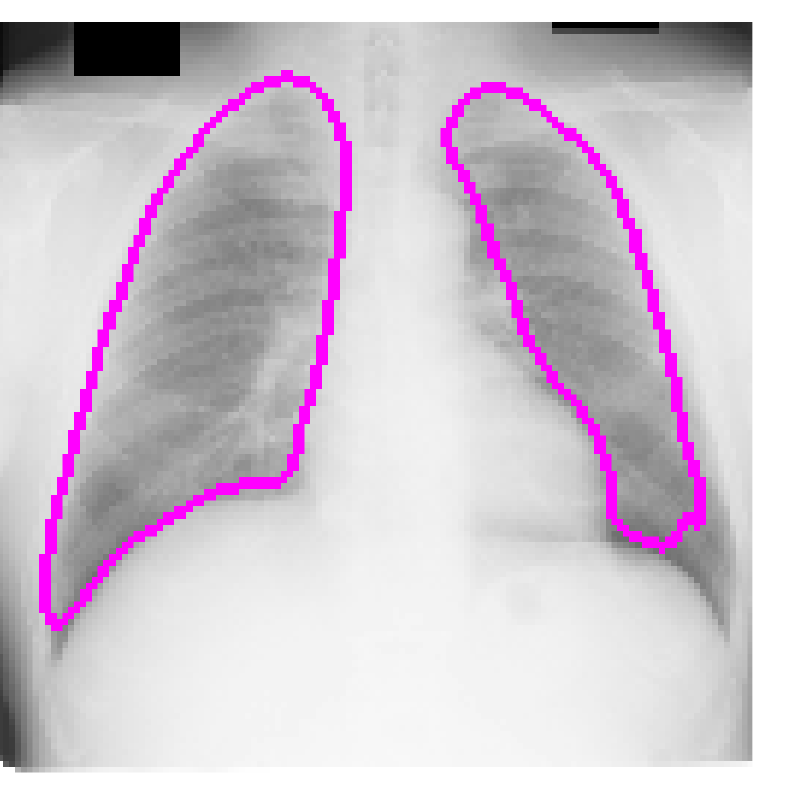}
    \\
    \noalign{\smallskip}
    {\Large \rotatebox{90}{U-MTL}} 
    &
    \includegraphics[width=0.2\linewidth, trim={4cm 1cm 3cm 1cm},clip]{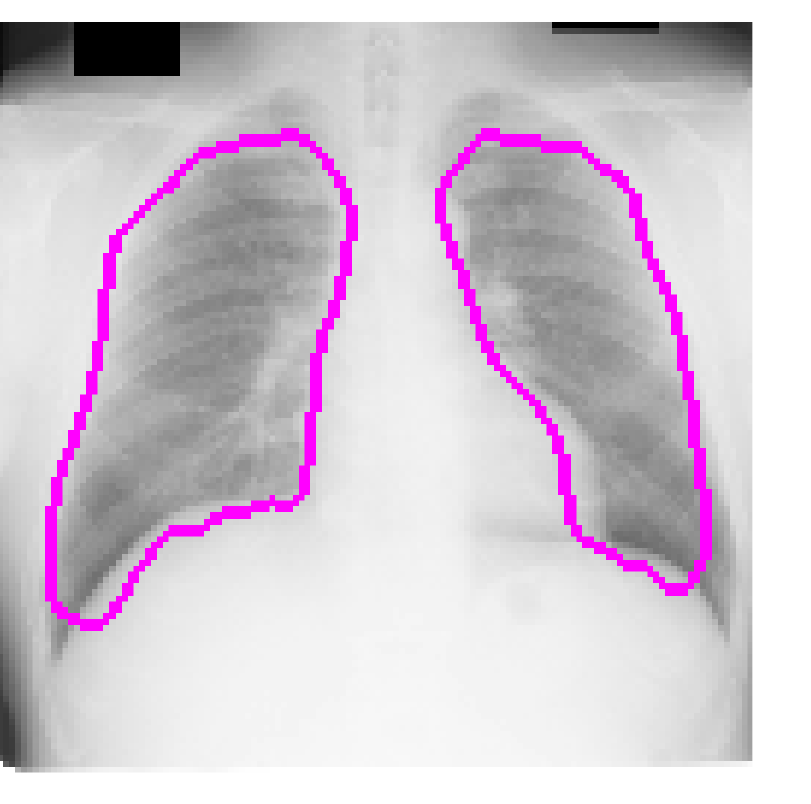}
    &
    \includegraphics[width=0.2\linewidth, trim={4cm 1cm 3cm 1cm},clip]{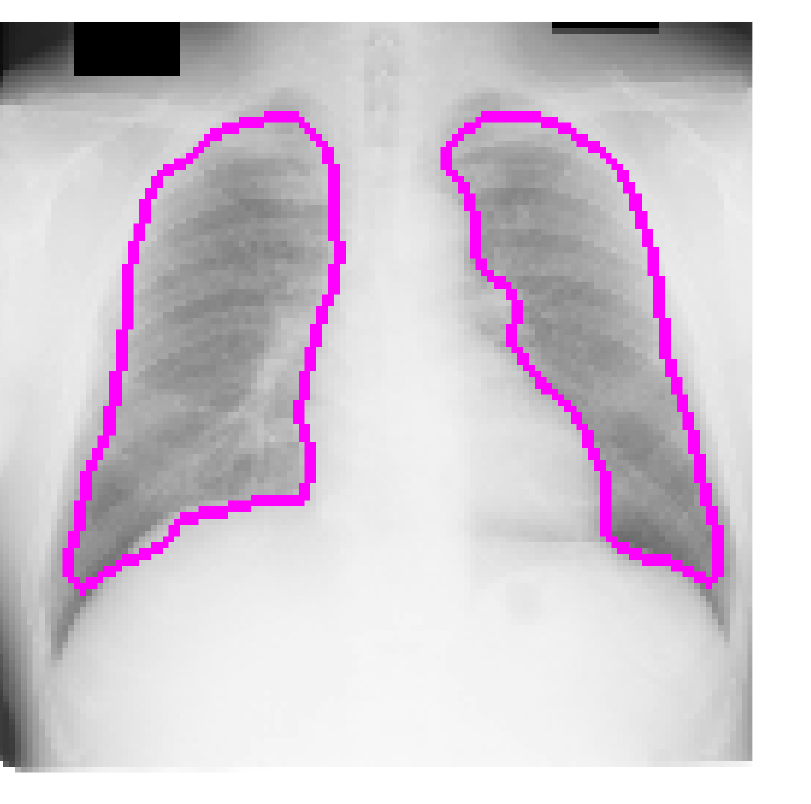}
    &
    \includegraphics[width=0.2\linewidth, trim={4cm 1cm 3cm 1cm},clip]{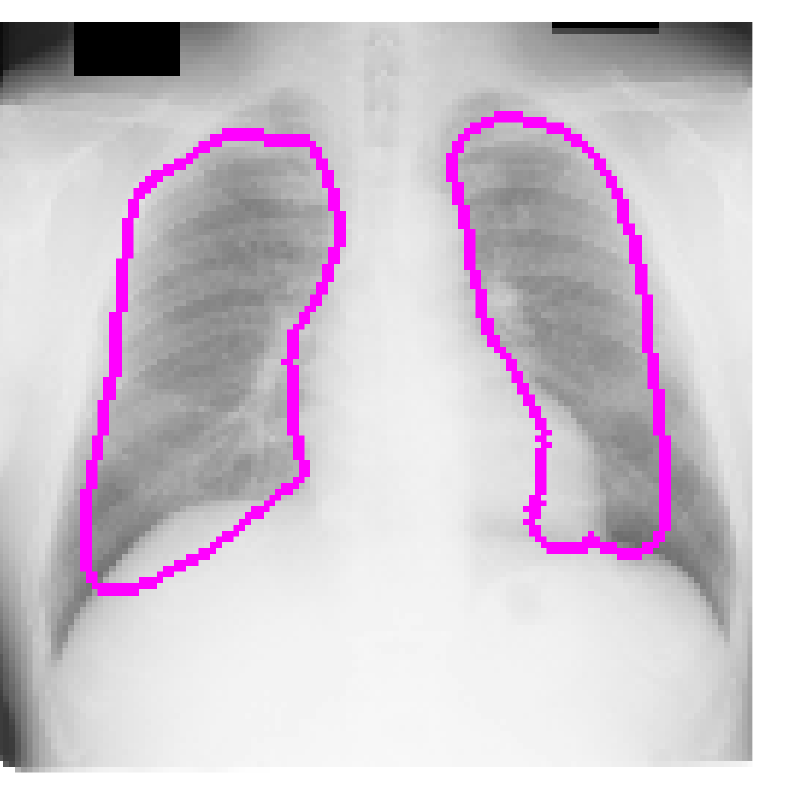}
    &
    \includegraphics[width=0.2\linewidth, trim={4cm 1cm 3cm 1cm},clip]{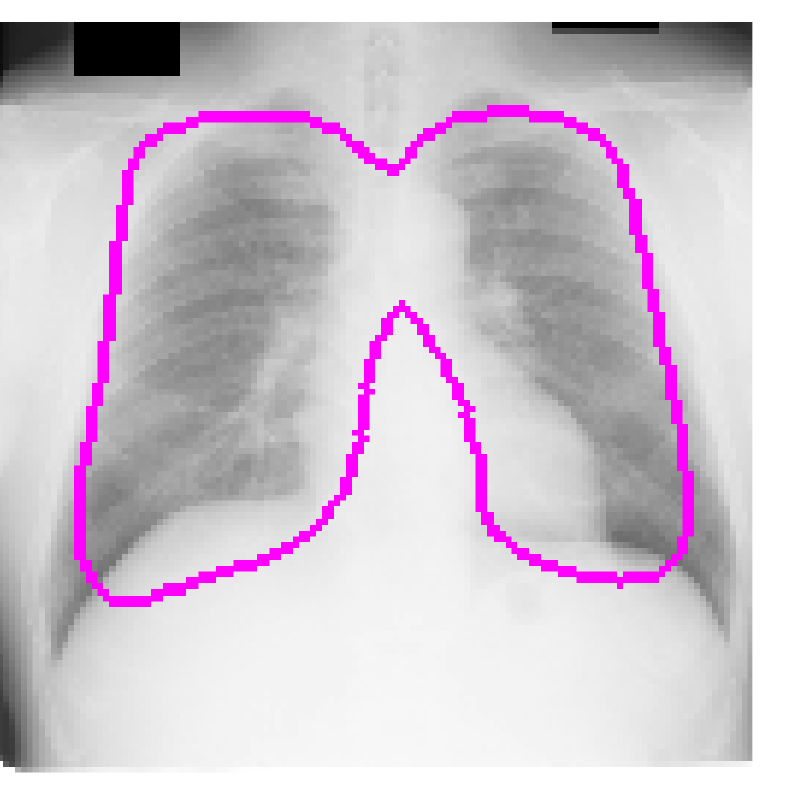}
    &
    \includegraphics[width=0.2\linewidth, trim={4cm 1cm 3cm 1cm},clip]{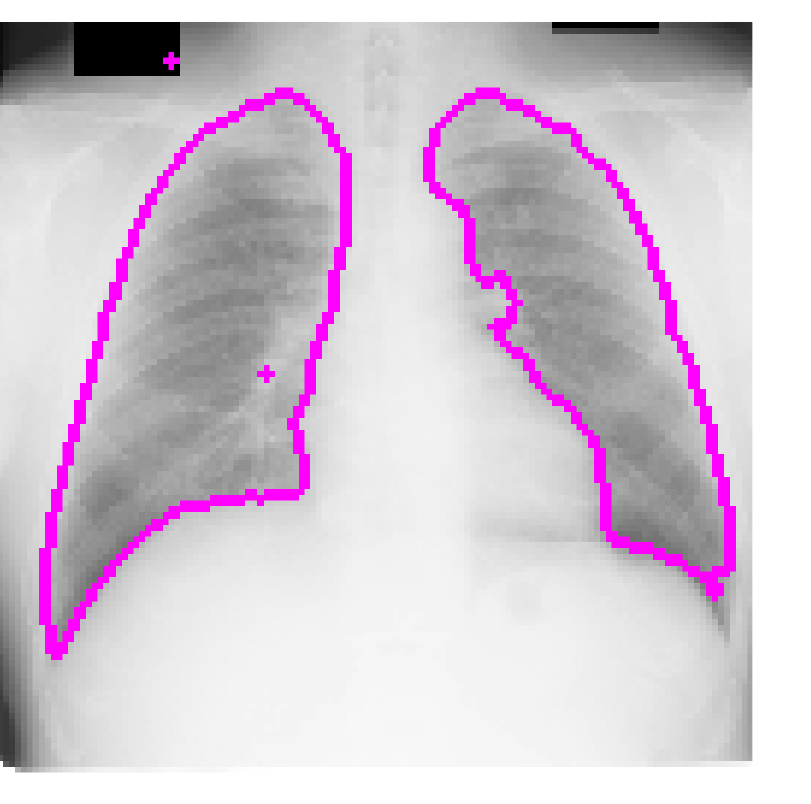}
    &
    \includegraphics[width=0.2\linewidth, trim={4cm 1cm 3cm 1cm},clip]{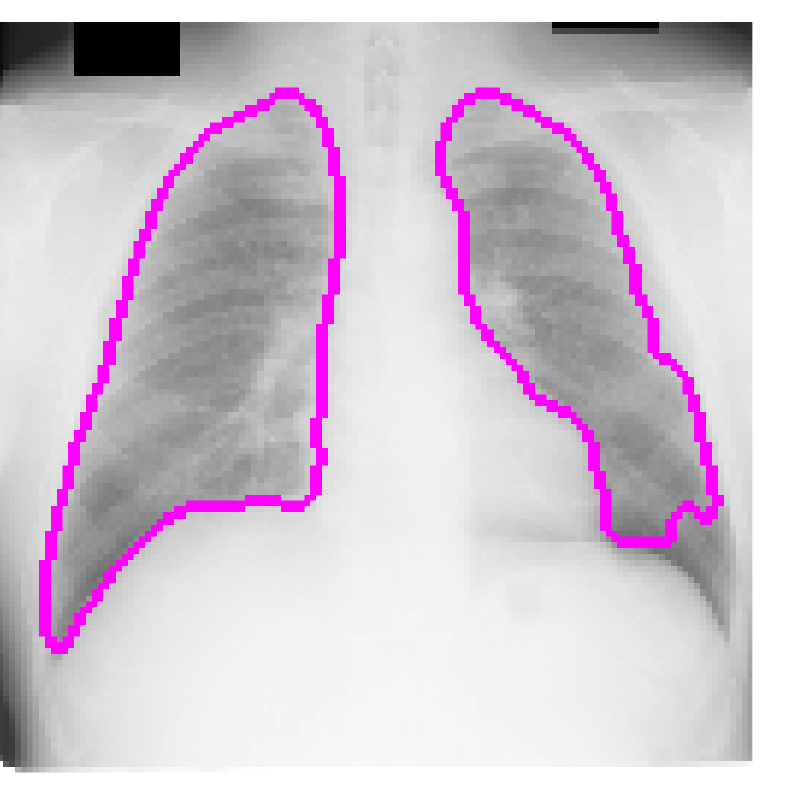}
    \\
    \noalign{\smallskip}
    {\Large \rotatebox{90}{S${}^2$MTL}} 
    &
    \includegraphics[width=0.2\linewidth, trim={4cm 1cm 3cm 1cm},clip]{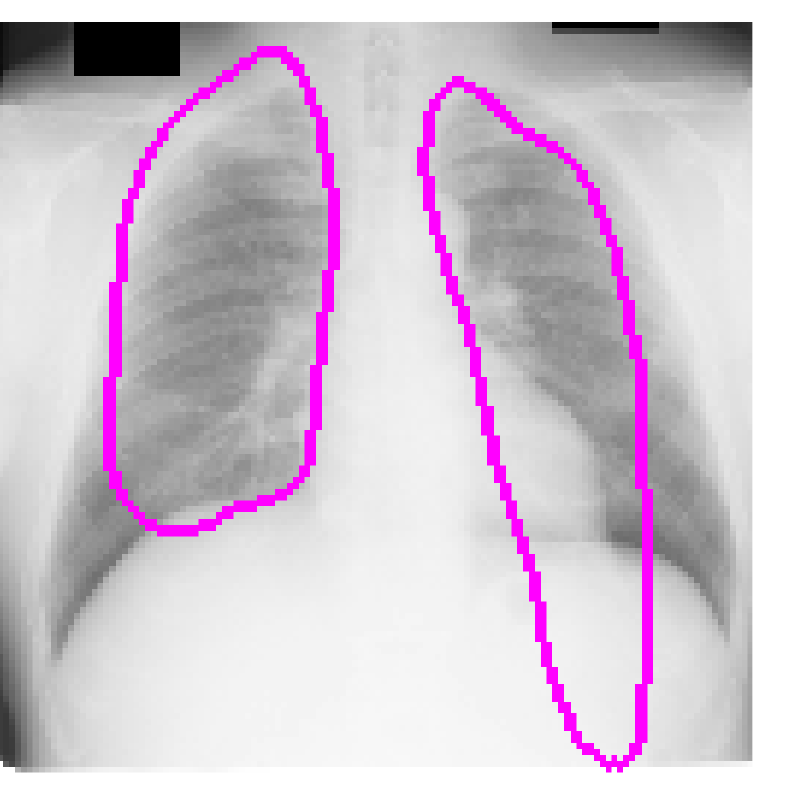}
    &
    \includegraphics[width=0.2\linewidth, trim={4cm 1cm 3cm 1cm},clip]{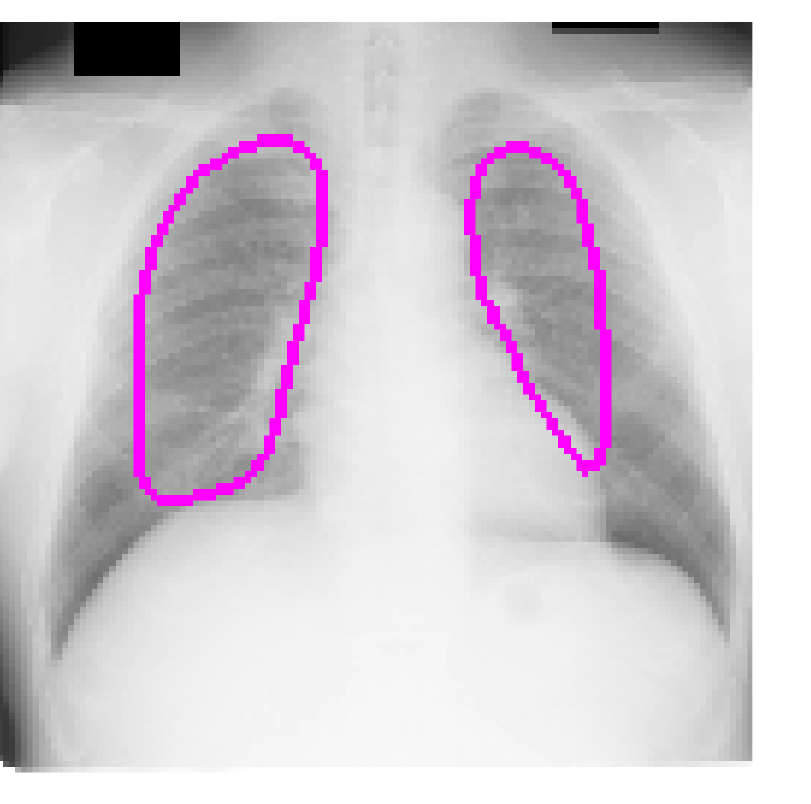}
    &
    \includegraphics[width=0.2\linewidth, trim={4cm 1cm 3cm 1cm},clip]{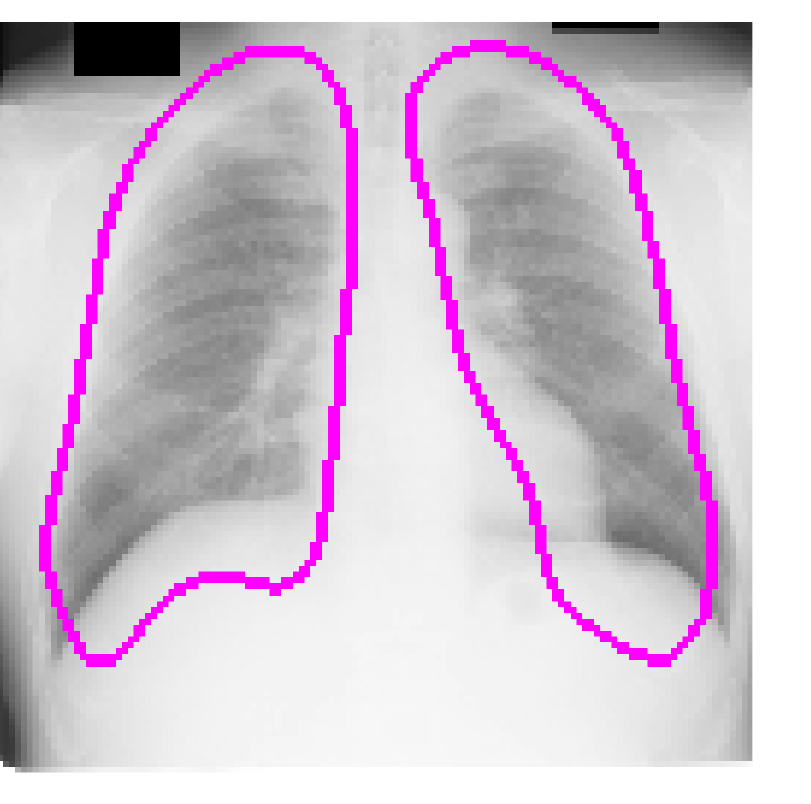}
    &
    \includegraphics[width=0.2\linewidth, trim={4cm 1cm 3cm 1cm},clip]{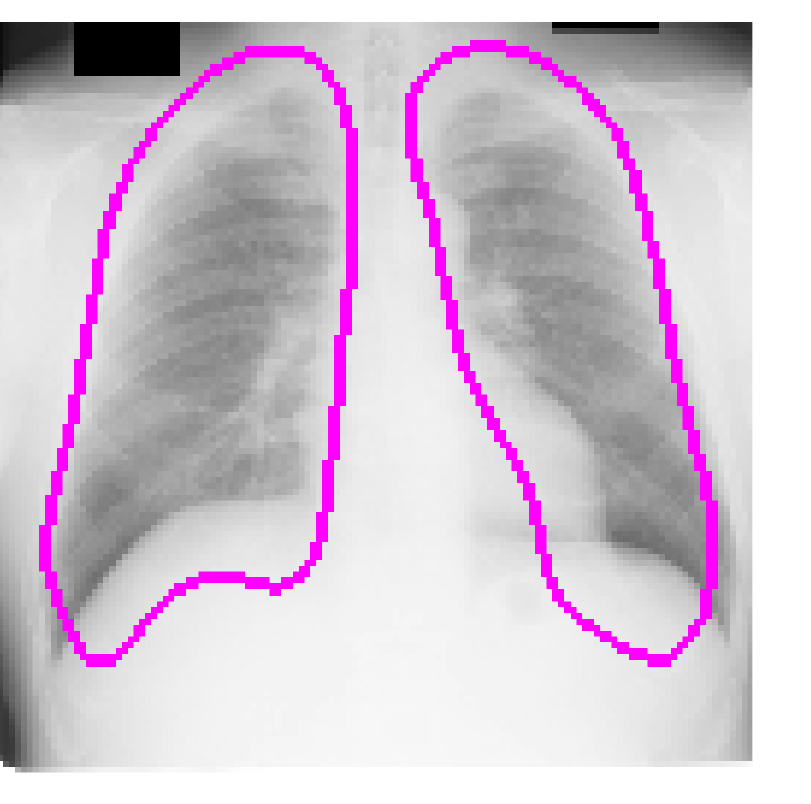}
    &
    \includegraphics[width=0.2\linewidth, trim={4cm 1cm 3cm 1cm},clip]{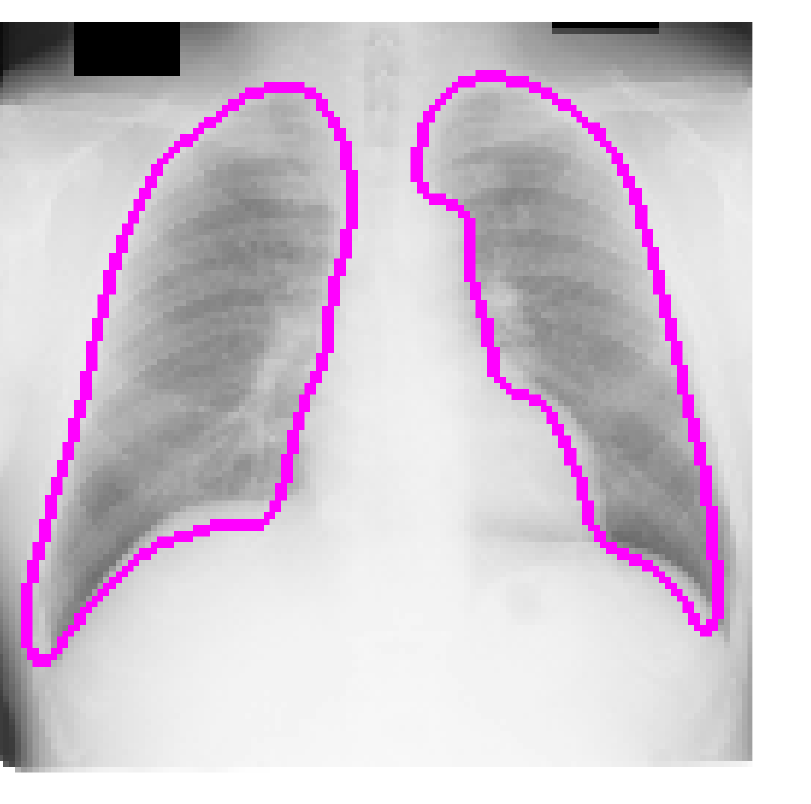}
    \\

    \noalign{\smallskip}
    {\Large \rotatebox{90}{S${}^4$MTL}} 
    &
    \includegraphics[width=0.2\linewidth, trim={4cm 1cm 3cm 1cm},clip]{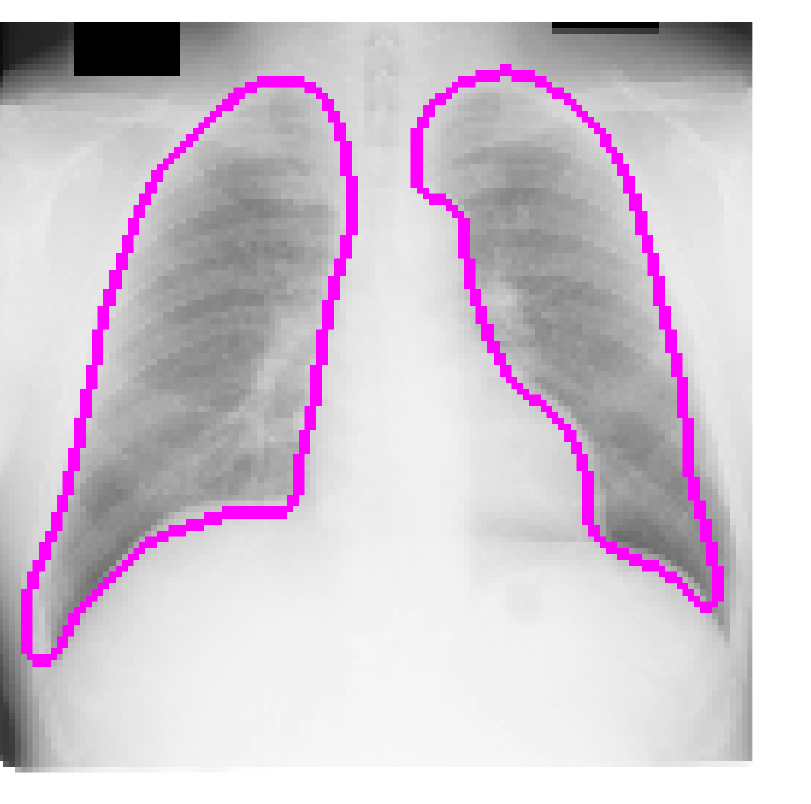}
    &
    \includegraphics[width=0.2\linewidth, trim={4cm 1cm 3cm 1cm},clip]{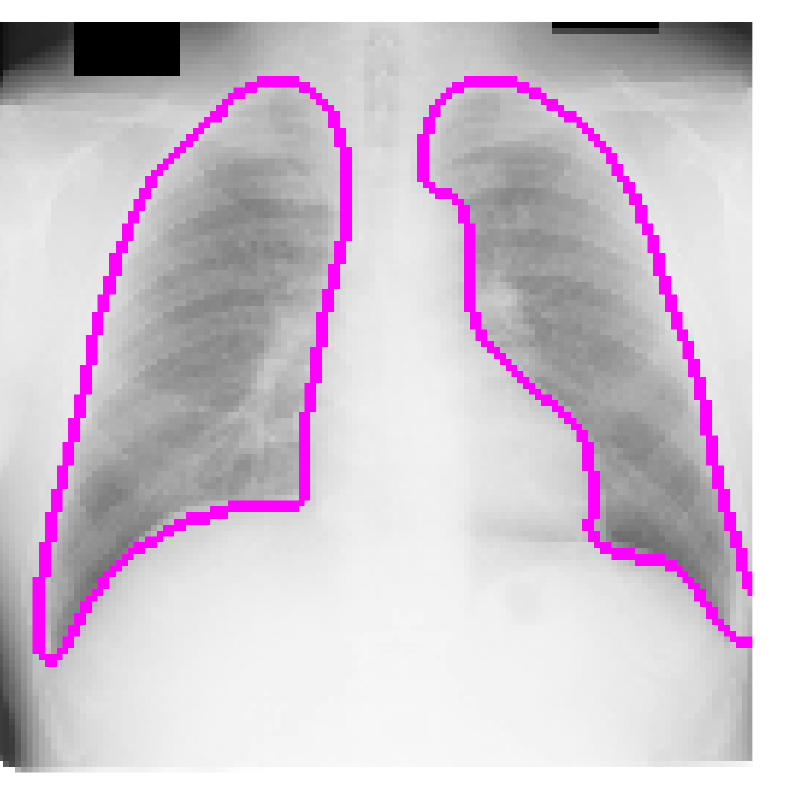}
    &
    \includegraphics[width=0.2\linewidth, trim={4cm 1cm 3cm 1cm},clip]{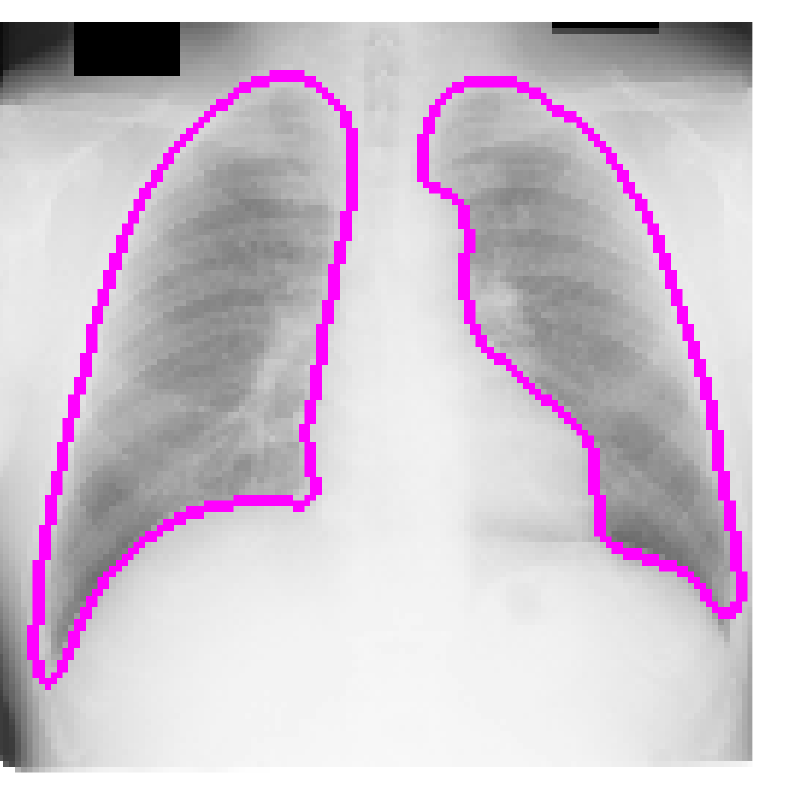}
    &
    \includegraphics[width=0.2\linewidth, trim={4cm 1cm 3cm 1cm},clip]{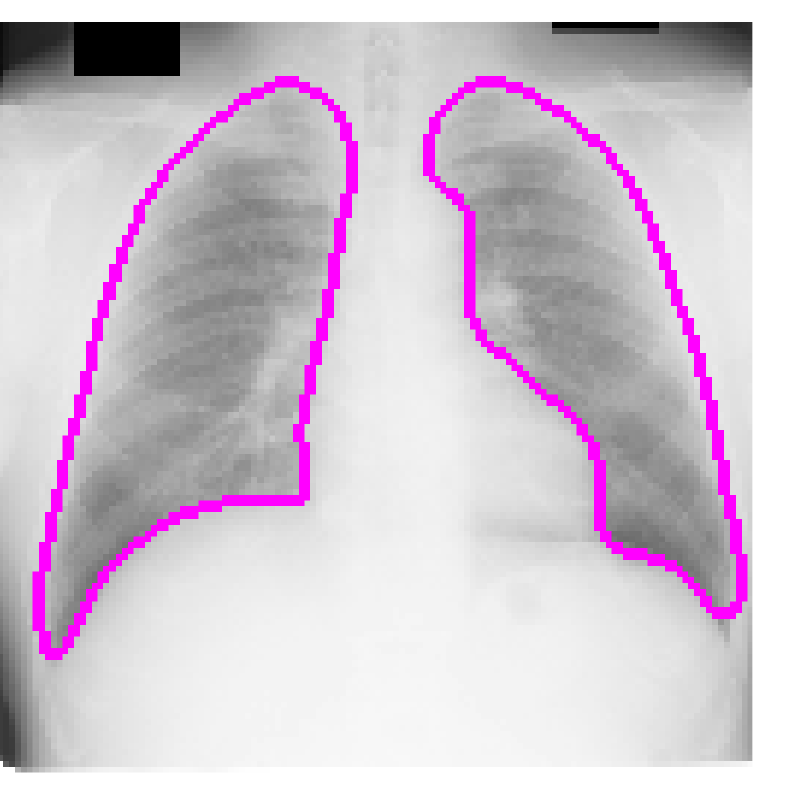}
    &
    \includegraphics[width=0.2\linewidth, trim={4cm 1cm 3cm 1cm},clip]{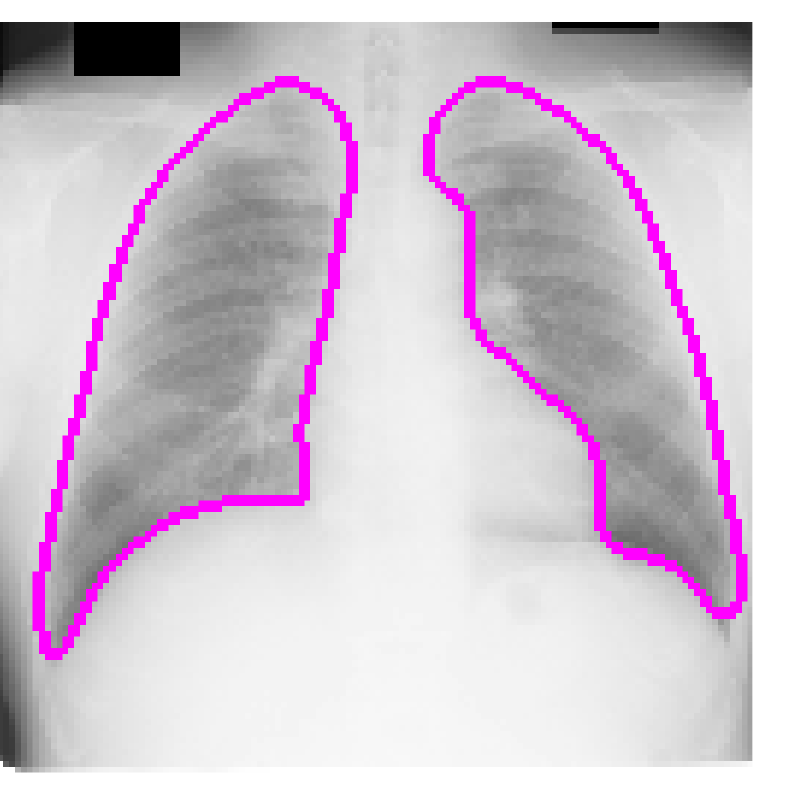}
    \\
    \end{tabular} }
    \caption{Chest Dataset: Boundary visualization of the predicted lung masks in a chest X-Ray shows consistent improvement with our S$^4$MTL model against the semi-supervised and fully-supervised baselines with varying proportions of labeled data.}
    \label{fig:chest_vis}
\end{figure}

\textbf{Segmentation:} The segmentation performance is evaluated both qualitatively and quantitatively. As shown in Table~\ref{table:seg-spine} and Table~\ref{table:seg-chest}, our S$^4$MTL model achieves the best scores compared to the semi-supervised, fully-supervised, single-task, and multitask models. The robustness of our S$^4$MTL model is confirmed by its consistent performance for different proportions of labeled data (Fig. \ref{fig:box_plots}). Visualization of the segmented vertebra and lung boundaries by different models in varying labeled data proportions (Fig.~\ref{fig:spine_vis} and Fig.~\ref{fig:chest_vis}) confirms the effectiveness of our model over competing models. For a fair comparison of all the models, a common segmentation backbone is used in the single-task for segmentation and in the multitask for segmentation mask generator. Our S$^4$MTL model consistently performs better than the semi/fully-supervised single-task (U-Net) and multitask (U-MTL) models, given the same number of labeled data during training ($|\mathcal{D_L}|$). The advantage really comes with the knowledge gain from the larger proportion of unlabeled data and multitask learning, S$^4$MTL's forte. This clearly reveals the effectiveness of our model.

\begin{table}[t]
\setlength{\tabcolsep}{4pt}
\centering
\caption{Spine performance: Vertebra segmentation comparison of the S${}^4$MTL model against the baselines for supervised/semi-supervised single-task and multitasking in different data settings with varying proportion of labeled data.}
\medskip
\label{table:seg-spine}
\resizebox{\linewidth}{!}{
\begin{tabular}{c c l c c c c c c}
            \toprule
           Dataset
           &
           Type
           &
          Model
          &
           DS
           &
           JI
           &
           SSIM
           &
           HD
           &
           Prec
           &
           Rec
           \\
           \midrule
           \multirow{24}{*}{\rotatebox{45}{Spine}}
           &
           \multirow{6}{*}{\rotatebox{70}{Single-Task}}
           &
           U-Net-100\% 
           &
           0.931 & 0.872&0.874& 4.335& 0.954&0.911 
           \\
           &&
           U-Net-50\% 
           &
           0.919 & 0.851&0.857&4.569& 0.949&0.893 
           \\
           &&
           U-Net-30\% 
           &
           0.910& 0.835&0.847&4.836& 0.946&0.877 
           \\
           &&
           U-Net-20\% 
           &
           0.903 & 0.824&0.839&5.022& 0.935&0.873 
           \\
           &&
           U-Net-10\% 
           &
           0.874 & 0.776&0.801& 5.276& 0.923&0.830 
           \\
           &&
           U-Net-5\% 
           &
           0.702 & 0.541&0.685&6.909& 0.957&0.554 
           \\
           \cmidrule{2-9}
           &
           \multirow{16}{*}{\rotatebox{70}{Multitask}}
           &
           U-MTL-100\%
           & 
           0.888 & 0.799 & 0.817& 5.348&0.908& 0.869  
           \\
           &&
           U-MTL-50\%
           & 
           0.890 & 0.802 & 0.827& 5.429&0.936& 0.849  
           \\
           &&
           U-MTL-30\%
           & 
           0.881 & 0.787 & 0.817& 5.421&0.950& 0.821  
           \\
           &&
           U-MTL-20\%
           & 
           0.862 & 0.758 & 0.792&5.733& 0.902&0.826  
           \\
           &&
           U-MTL-10\%
           & 
           0.873 & 0.775 & 0.804&6.396&0.884&0.863  
           \\
           &&
           U-MTL-5\%
           & 
           0.856 & 0.333 & 0.584&7.018& 0.999&0.333  
           \\
           \cmidrule{3-9}
           &&
           S$^2$MTL-50\%
           & 
           0.889 & 0.801 & 0.821& 4.956&0.887& 0.893  
           \\
           &&
           S$^2$MTL-30\%
           & 
           0.752 & 0.603 & 0.719& 6.032&0.997& 0.603  
           \\
           &&
           S$^2$MTL-20\%
           & 
           0.672 & 0.506 & 0.670&6.256& 0.998&0.506  
           \\
           &&
           S$^2$MTL-10\%
           & 
           0.640 & 0.471 & 0.654&6.772&0.998&0.471  
           \\
           &&
          S$^2$MTL-5\%
           & 
           0.500 & 0.333 & 0.584&7.018& 0.999&0.333  
           \\
           \cmidrule{3-9}
           &&
          S$^4$MTL-50\%
           & 
           {\bf0.934} & 0.876 & 0.875&3.762& 0.947&0.921  
           \\
           &&
           S$^4$MTL-30\%
           & 
           0.925 & 0.861 & 0.866&3.912& 0.946&0.905  
           \\
           &&
           S$^4$MTL-20\%
           & 
           0.921 & 0.853 & 0.860&4.162& 0.942&0.900  
           \\
           &&
           S$^4$MTL-10\%
           & 
           0.907 & 0.830 & 0.842&4.723&0.914&0.901  
           \\
           &&
          S$^4$MTL-5\%
           & 
           0.890 & 0.802 & 0.821&4.835& 0.887&0.893  
           \\
           \bottomrule
                \end{tabular}
}
\end{table}

\begin{figure}
    \centering
    \includegraphics[width=\linewidth, trim={0cm 0.25cm 0cm 0cm},clip]{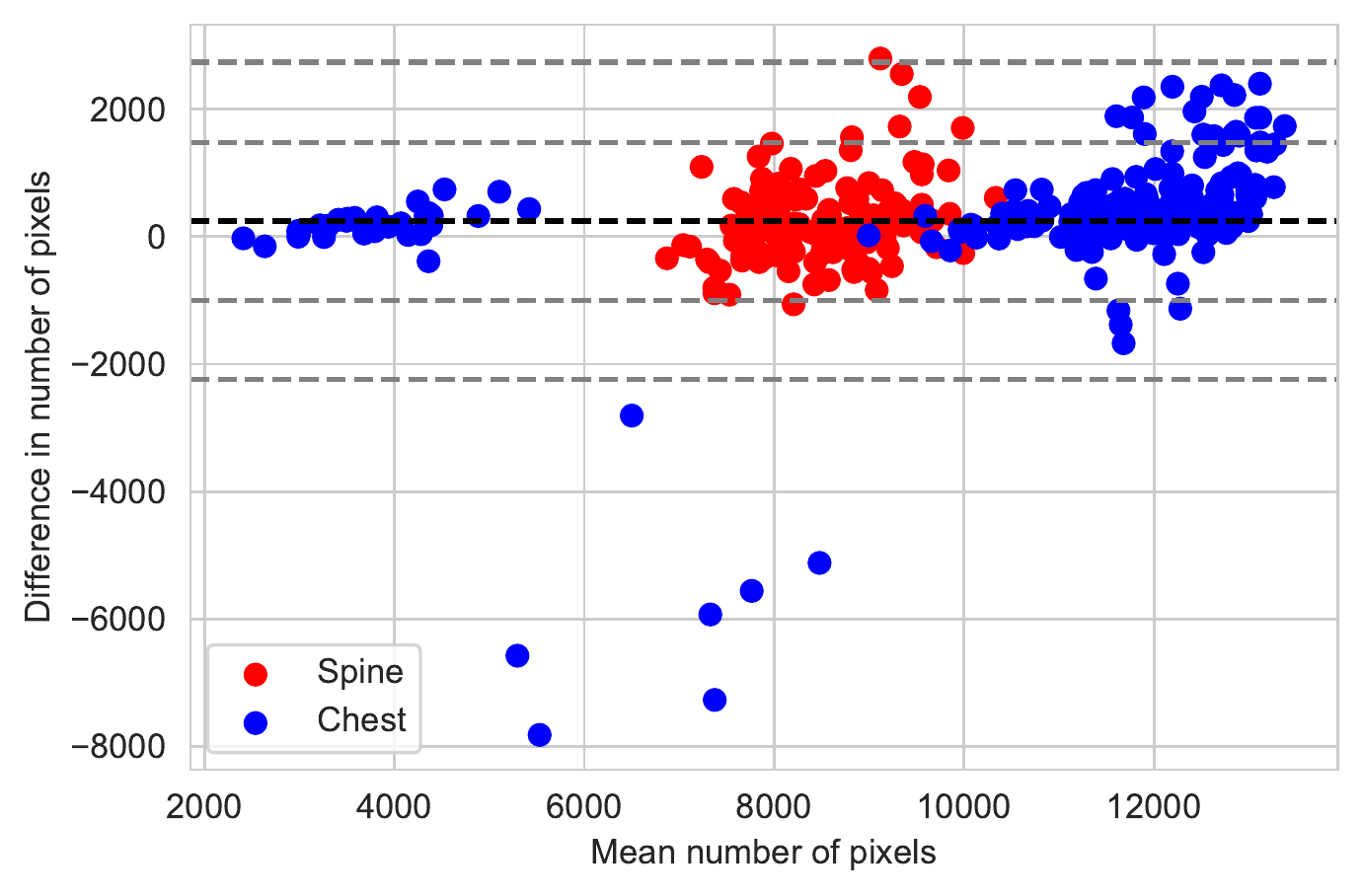}
    \caption{Bland-Altman plots show good agreement between the number of ground truth pixels and model (S$^4$MTL-50$\%)$ predicted pixels for the test sets.}
    \label{fig:bland-alt}
\end{figure}

\begin{figure}
    \centering
    \resizebox{\linewidth}{!}{
    \begin{tabular}{c c}
    \includegraphics[width=\linewidth]{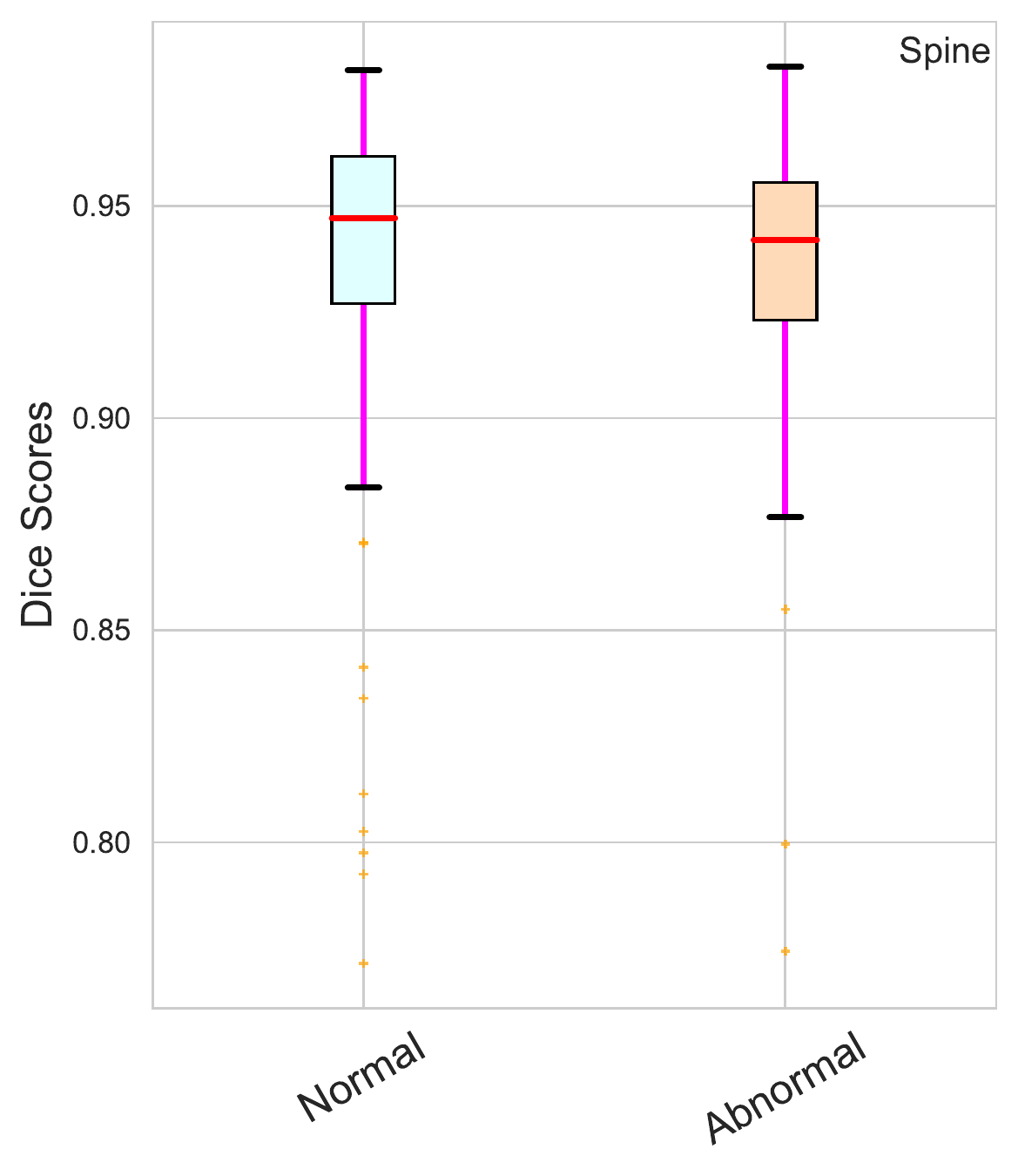}
    &
    \includegraphics[width=\linewidth]{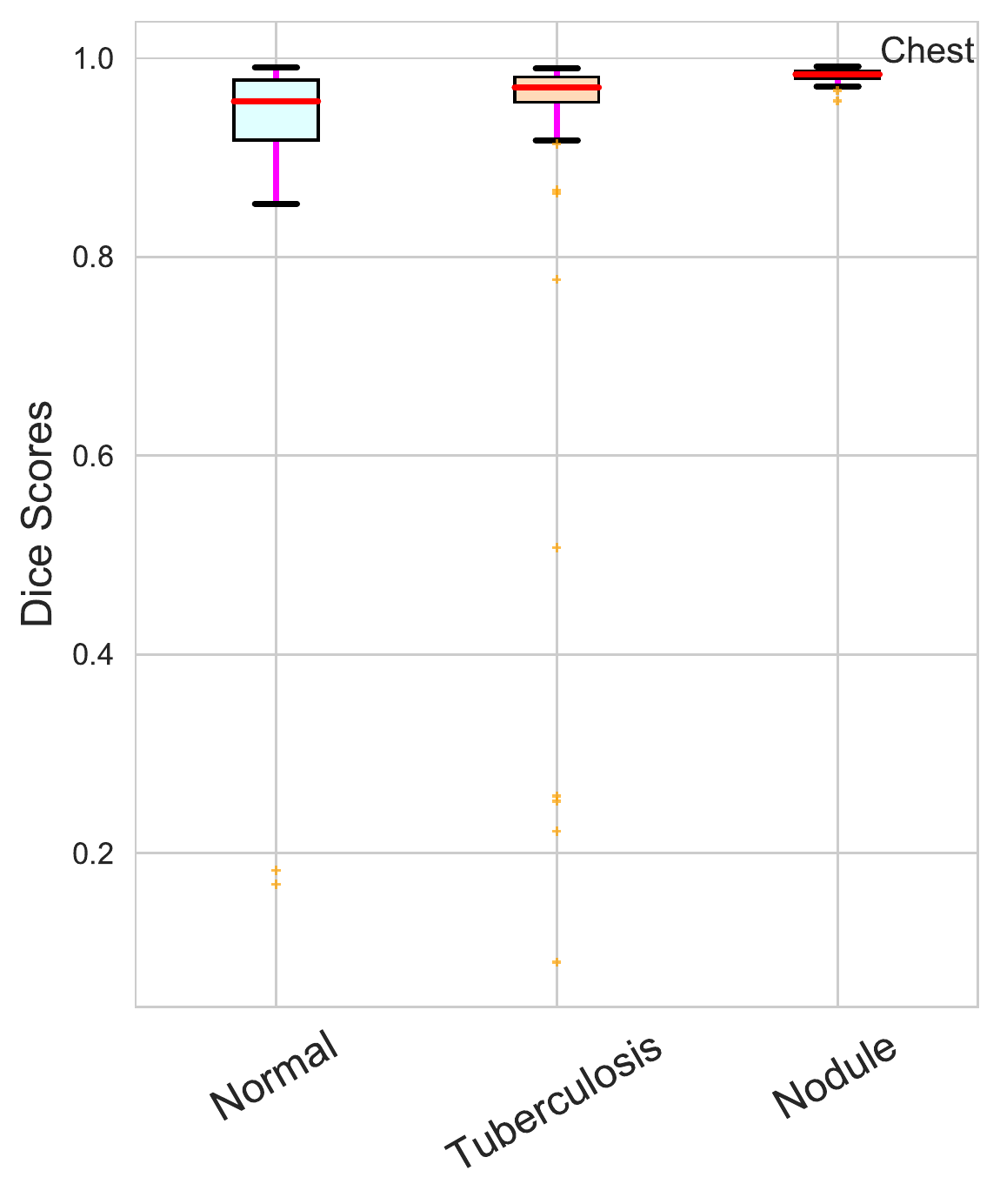}
    \end{tabular}
    }
    \caption{Class-wise boxplots showing the robustness of our S$^4$MTL-50$\%$ model in segmenting vertebrae and lungs from spinal and chest X-Rays.}
    \label{fig:class-boxs}
\end{figure}

\textbf{Statistical Analysis:} To verify the significance of our model, we perform three different statistical tests, namely one-way ANOVA, paired-t, and Wilcoxon signed-rank tests. All three tests confirm that our S$^4$MTL model with $50\%$ labeled data model shows significant improvement over all the semi-supervised and fully-supervised baselines in the Chest Dataset ($\textrm{p-value} < 0.05$). Moreover, for the Spine Dataset, our S$^4$MTL model with $50\%$ labeled data is found to be significantly different from all the baseline models except the fully-supervised U-Net model. The Bland-Altman plots shown in Fig.~\ref{fig:bland-alt} also suggest good agreement between ground truth and the prediction by our S$^4$MTL model for both datasets. Moreover, we perform a robustness analysis to compare the class-wise segmentation performance of our S$^4$MTL model. Fig.~\ref{fig:class-boxs} shows that there are no significant differences among the class-wise segmentation performances (Dice scores). Furthermore, by performing one-way ANOVA, independent sample t-test, and Pearson correlation, we confirm that our model is robust against different disease classes (normal vs abnormal in the Spine dataset and normal vs TB vs nodule in the Chest dataset).

\begin{table}
\setlength{\tabcolsep}{4pt}
\centering
\caption{Chest performance: Lung segmentation comparison of the S${}^4$MTL model against the baselines for semi-supervised multitasking in different data settings with varying proportion of labeled data.}
\medskip
\label{table:seg-chest}
\resizebox{\linewidth}{!}{
\begin{tabular}{c c l c c c c c c}
            \toprule
           Dataset
           &
           Type
           &
          Model
          &
           DS
           &
           JI
           &
           SSIM
           &
           HD
           &
           Prec
           &
           Rec
           \\
           \midrule
           \multirow{24}{*}{\rotatebox{45}{Chest}}
           &
           \multirow{6}{*}{\rotatebox{70}{Single-Task}}
           &
           U-Net-100\% 
           &
           0.922 & 0.856&0.816&4.524&0.909&0.936 
           \\
           &&
           U-Net-50\% 
           &
           0.892 & 0.806&0.761&5.382& 0.846&0.945 
           \\
           &&
           U-Net-30\% 
           &
           0.834 & 0.715&0.654&6.584& 0.723&0.986 
           \\
           &&
           U-Net-20\% 
           &
           0.815 & 0.688&0.636&5.979& 0.758&0.882 
           \\
           &&
           U-Net-10\% 
           &
           0.800 & 0.667&0.599&7.111& 0.682&0.969 
           \\
           &&
           U-Net-5\% 
           &
           0.788 & 0.650&0.583&7.419& 0.654&0.991 
           \\
           \cmidrule{2-9}
           &
           \multirow{16}{*}{\rotatebox{70}{Multitask}}
           &
           U-MTL-100\%
           & 
           0.874 & 0.776 & 0.747& 4.969&0.800& 0.963 
           \\
           &&
           U-MTL-50\%
           & 
           0.888 & 0.799 & 0.756& 4.700&0.822& 0.965 
           \\
           &&
           U-MTL-30\%
           & 
           0.820&0.694&0.658&5.374&0.766& 0.881  
           \\
           &&
           U-MTL-20\%
           & 
           0.829&0.708&0.654& 6.483&0.719& 0.979  
           \\
           &&
           U-MTL-10\%
           & 
           0.841 & 0.726 & 0.683& 5.554&0.746&0.964  
           \\
           &&
           U-MTL-5\%
           & 
           0.820 & 0.695 & 0.644& 5.648&0.748& 0.908  
           \\
           \cmidrule{3-9}
           &&
           S$^2$MTL-50\%
           & 
           0.810 & 0.681 & 0.636& 4.862&0.782& 0.840 
           \\
           &&
           S$^2$MTL-30\%
           & 
           0.572&0.401&0.465&5.416&0.824& 0.438  
           \\
           &&
           S$^2$MTL-20\%
           & 
           0.607&0.436&0.516& 5.782&0.987& 0.438  
           \\
           &&
           S$^2$MTL-10\%
           & 
           0.558 & 0.387 & 0.487& 5.827&0.998&0.387  
           \\
           &&
          S$^2$MTL-5\%
           & 
           0.458 & 0.300 & 0.421& 7.496&0.989& 0.300  
           \\
           \cmidrule{3-9}
          &&
          S$^4$MTL-50\%
           & 
           {\bf0.946} & 0.898 & 0.864&3.432&0.964&0.939  
           \\
           &&
           S$^4$MTL-30\%
           & 
           0.903 & 0.823 & 0.798& 3.963& 0.885&0.921  
           \\
           &&
           S$^4$MTL-20\%
           & 
           0.895 & 0.810 & 0.784&4.119& 0.878&0.912
           \\
           &&
           S$^4$MTL-10\%
           & 
           0.868 & 0.768 & 0.742&4.272& 0.834&0.905  
           \\
           &&
          S$^4$MTL-5\%
           & 
           0.845 & 0.732 & 0.703& 4.619& 0.811&0.883  
           \\
           \bottomrule
        \end{tabular}
}
\end{table}

\section{Conclusions}

Learning from small labeled datasets has been one of the most challenging tasks in computer vision and medical imaging. We have proposed a novel self-supervised, semi-supervised, multitask learning (S$^4$MTL) model and validated it in medical image classification and segmentation experiments with limited labeled data. The effectiveness of our S$^4$MTL model over semi-supervised and fully-supervised single-task and multitask models is confirmed by the experimental results. Moreover, S$^4$MTL with 50\% labeled data achieves better performance, with statistical significance, over all other semi-supervised models. 

Our model may further, be utilized in general settings when data available with and without labels have dissimilar distributions. A worthwhile next step would be experimenting with more sophisticated medical imaging and computer vision data at larger scale and in more challenging settings.

\bibliography{example_paper}
\bibliographystyle{icml2020}

\end{document}